\documentclass[10pt,twocolumn,letterpaper]{article}

\usepackage[pagenumbers]{cvpr} %

\usepackage[pass]{geometry}

\definecolor{darkgreen}{rgb}{0,0.7,0}
\definecolor{darkblue}{RGB}{31,119,180}
\definecolor{darkred}{RGB}{214,39,40}

\usepackage{tikz}
\usetikzlibrary{shapes}
\usetikzlibrary{calc}%

\usepackage[T1]{fontenc}
\usepackage[utf8]{inputenc}

\usepackage{multirow}

\usepackage{bbding}

\definecolor{cvprblue}{rgb}{0.21,0.49,0.74}
\usepackage[pagebackref,breaklinks,colorlinks,allcolors=cvprblue]{hyperref}

\def\paperID{} %
\def\confName{CVPR}
\def\confYear{2026}

\title{\benchName{}: Benchmarking Closed-Loop Driving Generalization}

\author{
Simon Gerstenecker \qquad
Andreas Geiger \qquad
Katrin Renz \\
[4mm]
University of Tübingen \quad
Tübingen AI Center \quad \\
}

\usepackage{siunitx}

\newcommand{\boldparagraph}[1]{\vspace{0.2cm}\noindent{\textbf{#1.}} } %

\newcommand{\benchName}{Fail2Drive}
\newcommand{\supp}{supplementary material}

\usepackage{placeins}

\begin{document}
\twocolumn[{%
\renewcommand\twocolumn[1][]{#1}%
\vspace{-14.0mm}
\maketitle
\begin{center}
    \centering
    \captionsetup{type=figure}
    \includegraphics[width=0.99\textwidth]{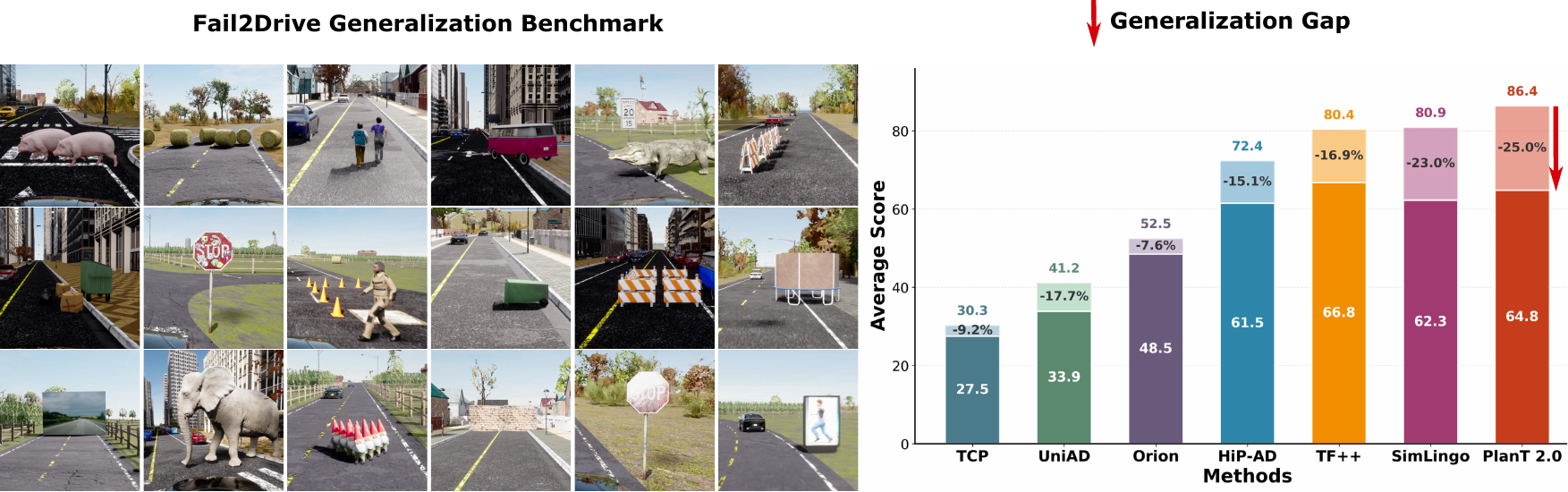} \vspace{-2.0mm}
    \captionof{figure}{\textbf{Overview:} Fail2Drive introduces the first paired-route benchmark for closed-loop generalization on truly unseen long-tail scenarios in CARLA. It turns qualitative failures into measurable generalization gaps. Evaluating seven recent driving models exposes strong shortcut learning and missing fallback behavior, revealing where current approaches break and where progress is most needed.}
\label{fig:teaser}

\end{center}}]
\begin{abstract}

Generalization under distribution shift remains a central bottleneck for closed-loop autonomous driving. Although simulators like CARLA enable safe and scalable testing, existing benchmarks rarely measure true generalization: they typically reuse training scenarios at test time. Success can therefore reflect memorization rather than robust driving behavior. We introduce Fail2Drive, the first paired-route benchmark for closed-loop generalization in CARLA, with 200 routes and 17 new scenario classes spanning appearance, layout, behavioral, and robustness shifts. Each shifted route is matched with an in-distribution counterpart, isolating the effect of the shift and turning qualitative failures into quantitative diagnostics. Evaluating multiple state-of-the-art models reveals consistent degradation, with an average success-rate drop of 22.8\%. Our analysis uncovers unexpected failure modes, such as ignoring objects clearly visible in the LiDAR and failing to learn the fundamental concepts of free and occupied space.
To accelerate follow-up work, Fail2Drive includes an open-source toolbox for creating new scenarios and validating solvability via a privileged expert policy. Together, these components establish a reproducible foundation for benchmarking and improving closed-loop driving generalization. We open-source all code, data, and tools at \url{https://github.com/autonomousvision/fail2drive}.

\end{abstract}

\section{Introduction}
\label{sec:intro}

Autonomous driving has seen remarkable progress over the last years, transitioning from modular pipelines~\cite{Thorpe1988PAMI,  Xu2014ICRA, Cui2021ICCV} 
to end-to-end~\cite{chen2023end, Chitta2023PAMI, Codevilla2019ICCV}
and vision-language-action (VLA) models~\cite{Fu2025Orion, Zhou2025AutoVLA, Renz2025cvpr} that promise to learn driving behavior directly from data at scale, showing promising performance gains.
Yet, a central question remains open: \emph{do these models actually generalize to rare unseen situations}?

Simulators are a compelling option to find an answer. In theory, they make it possible to instantiate rare, safety-critical, or legally problematic situations without endangering anyone, and they provide standardized interfaces so that different algorithms can be compared on equal grounds. CARLA~\cite{CarlaDosovitsky}, in particular, has emerged as the de facto standard simulator for closed-loop driving in driving research and has enabled a long line of driving work. However, the scenarios shipped with CARLA and used in the prominent CARLA-based benchmarks~\cite{jia2024bench,Leaderboard2024} are still limited in variability: they reuse a small set of assets, and are typically evaluated under protocols where models are trained on the same scenarios on which they are later tested. 
As a consequence, it is currently hard to test closed-loop out-of-distribution generalization.

At the same time, current Large-Language-Model (LLM)- and Reinforcement-Learning (RL)-based driving models~\cite{JaegerCarl2025, gigaflow, Renz2025cvpr, Fu2025Orion} make increasingly strong claims about robustness and generalization. To verify these claims, we need benchmarks that probe such out-of-distribution situations and can faithfully measure the generalization gap.

We introduce \textbf{Fail2Drive}, a benchmark and toolbox for evaluating the \emph{generalization capability} of autonomous driving models in CARLA. Fail2Drive instantiates a large set of new situations not included in the standard CARLA releases, e.g., obstacles with unseen assets, wild animals crossing the street, parked vehicles in unexpected positions, or visually adversarial scenarios.
Our key idea is to measure the \emph{generalization drop}, rather than relying solely on absolute performance scores. To this end, we introduce paired in-distribution and generalization routes in which location and traffic conditions are held constant, and only the targeted shift is varied. This paired design isolates the causal factor of failure and enables controlled %
evaluation across a spectrum of capabilities, ranging from perceptual robustness to downstream behavioral adaptation.
Because the route context is fixed, performance differences directly quantify sensitivity to the induced shift. This enables analysis of \emph{counterfactual scenarios}: e.g., would a policy still yield if a pedestrian were replaced by an animal, or still avoid a blockage if the obstacle asset were swapped while preserving the surrounding traffic and road geometry?
Importantly, unlike prior perception-focused failure tests that evaluate intermediate detection or classification outputs, our framework derives scores directly from planning performance. As a result, measured failures reflect end-to-end decision-making degradation rather than isolated perception errors.

In addition, Fail2Drive includes a toolbox layer on top of CARLA to make designing such scenarios less time-consuming and more user-friendly. 

Finally, we conduct an in-depth analysis of seven recent models on Fail2Drive, exposing significant and consistent failure modes. Examples include TF++ disregarding LiDAR information, colliding with obstacles blocking the ego path, a lack of fallback behavior in uncertain situations, e.g., when pedestrians walk on the road, and all models failing to learn a generalizable internal representation of obstacles.

\textbf{Contributions.} Our work makes the following contributions: (i) we propose \emph{Fail2Drive}, a closed-loop generalization benchmark, covering a wide range of new scenario classes across four categories. Fail2Drive uses paired in-distribution/generalization routes to directly quantify closed-loop generalization gaps. (ii) We provide a cross-paradigm evaluation of seven recent driving models, showing that current methods still rely on dataset- and simulator-specific regularities and that Fail2Drive exposes these brittle behaviors in a reproducible way; and (iii) we release a toolbox to construct and validate new scenario classes, extend the benchmark, and generate more diverse datasets with lower engineering overhead.

\section{Related Work}
\label{sec:relatedwork}

\boldparagraph{General Autonomous Driving Benchmarks}
Large-scale datasets have been central to the progress in autonomous driving. nuScenes~\cite{caesar2019nuscenes} provides standardized perception benchmarks, but its open-loop nature offers limited insight into planning quality~\cite{Codevilla2018ECCV, jaeger2024github}. nuPlan~\cite{Caesar2021CVPRW} introduces closed-loop planner evaluation using realistic traffic data, yet lacks diverse long-tail events. interPlan~\cite{Hallgarten2024ARXIV} adds safety-critical interactions but evaluates only planner outputs without sensor input. NAVSIM~\cite{Dauner2024NEURIPS} bridges this gap by enabling efficient sensor-based evaluation.

In addition, CARLA~\cite{Dosovitskiy2017CORL} has become the dominant simulator for closed-loop research, supporting multimodal sensors, reactive traffic, and diverse environments. 
Many benchmarks have been proposed~\cite{Codevilla2019ICCV, Chen2022CVPRa, Chitta2021ICCV, Chitta2023PAMI, jia2024bench}. 
These efforts have significantly advanced end-to-end driving~\cite{Zimmerlin2024tfpp, Renz2025cvpr, tang2025hipadhierarchicalmultigranularityplanning}, yet they reuse the same set of assets, actors, and scenario configurations as seen during training. 
As a result, they are unable to isolate whether models learn robust concepts or only memorize CARLA-specific patterns, making true out-of-distribution generalization difficult to assess.

\boldparagraph{Out-of-Distribution Evaluation in Driving}
Another line of work studies robustness under distribution shift. 
Perception-focused analyses explore shifts in appearance, occlusion, or adversarial perturbations~\cite{Suryanto2022cvprDTA, Nesti_2022_WACV, Maag2023_wildlife}. 
Other works generate safety-critical or adversarial scenarios~\cite{NEURIPS2022_safebench, zhang2024chatscene}; however, they produce mostly simplistic situations that are overly focused on the behavior of other actors and are partially already included in the basic CARLA scenarios. CARLA Real Traffic Scenarios~\cite{osiński2021carlarealtrafficscenarios} replays real-world traffic logs, and~\cite{9552860} proposes off-road scenes.
PlanT~2.0~\cite{gerstenecker2025plant2} reveals generalization failures in open-loop privileged planning, but the transferability to closed-loop sensor-based driving remains unclear. 
Another promising direction are NerF- or Gaussian-Splatting-based methods to obtain controllable simulators~\cite{ljungbergh2024neuroncap, zhou2024hugsim, lu2024urbancad}. 
However, existing works do not provide a high variety of visually and behaviorally out-of-distribution scenarios that can be tested closed-loop and lack causal attribution of the failure case. 

\section{\benchName{} - Generalization Benchmark}
\label{sec:method}

\benchName{} enables controlled measurement of the generalization gap of closed-loop driving models. It extends CARLA 0.9.15 with new scenarios while preserving full compatibility with existing driving stacks, requiring no architectural changes or custom integration.

\benchName{} is built around three principles: (i) \textbf{Distribution shift.} Routes include visual, geometric, and behavioral variations that differ from standard CARLA assets and layouts. (ii) \textbf{Paired evaluation.} Each generalization route has a corresponding in-distribution route representing an equivalent traffic situation at the same location without the shift. The paired route isolates robustness under shift. (iii) \textbf{Full extensibility.} All scenarios, assets, and behaviors can be authored and modified without editing CARLA core code, enabling reproducibility and community extension. 

Unlike prior work, \benchName{} does not only report a single driving score. Its design explicitly reveals which kinds of distribution shifts cause performance degradation.

\begin{figure}
    \centering
    \includegraphics[width=0.92\linewidth]{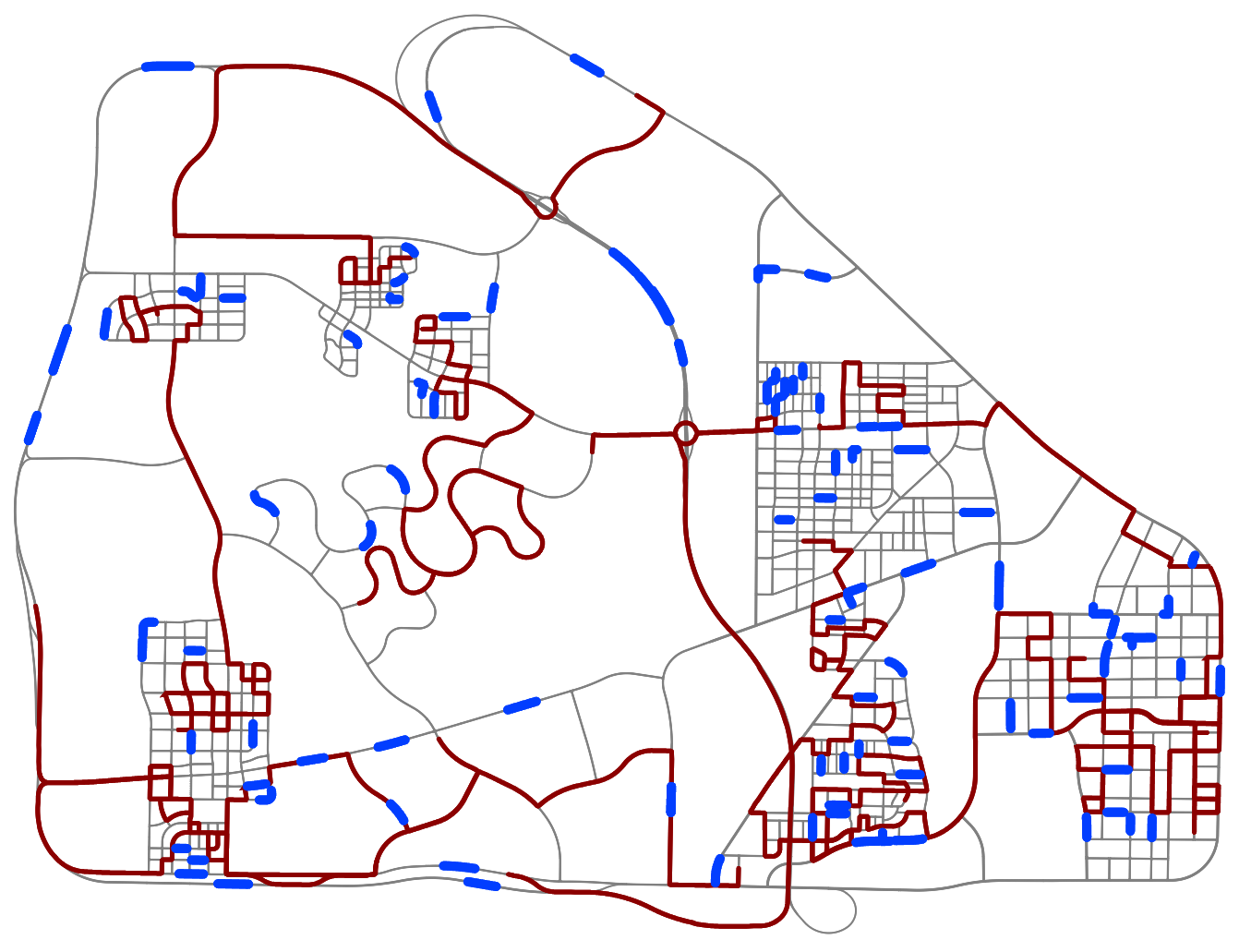}\vspace{-2.0mm}
    \caption{\textbf{Route diversity}. Fail2Drive routes (blue) are diversely spread across Town13, covering a wide range of environments, and have little overlap with the official CARLA validation routes (red).}
    \vspace{-4mm}
    \label{fig:diversity}
\end{figure}

\subsection{Benchmark Design}

\benchName{} contains 200 evaluation routes sampled across CARLAs validation Town 13, a large \SI{100}{\square\kilo\meter} map that includes residential neighborhoods, industrial districts, and rural segments (Figure~\ref{fig:diversity}). It covers a wide range of road widths, curvature, and speed limits. The routes were selected to have minimal overlap with the official CARLA validation routes, ensuring independence from prior benchmarks. All routes are short (average of 219 meters), following the widely used Bench2Drive benchmark, enabling clear failure attribution to a single scenario.

\begin{table*}[ht]
\centering
\scalebox{0.85}{
\begin{tabular}{l|cccc|c|ccc|ccc}
\toprule
\multirow{2}{*}{\textbf{Method}\vspace{-1mm}} 
& \multirow{2}{*}{\rotatebox{90}{RGB}}
& \multirow{2}{*}{\rotatebox{90}{LiDAR}}
& \multirow{2}{*}{\rotatebox{90}{Privileged}}
& \multirow{2}{*}{\rotatebox{90}{Learned}}
& \textbf{Bench2Drive} 
& \multicolumn{6}{c}{\textbf{Fail2Drive}}\\
\cmidrule(lr){6-12}
& & & &
& 
& \multicolumn{3}{c}{In-Distribution}\vline
& \multicolumn{3}{c}{Generalization} \\
\cmidrule(lr){7-12}
& & & &
& DS $\uparrow$
& DS $\uparrow$ & SR(\%) $\uparrow$ & HM $\uparrow$
& DS $\uparrow$ & SR(\%) $\uparrow$ & HM $\uparrow$\\
\midrule
TCP~\cite{Wu2022NeurIPS}  & \Checkmark & \XSolidBrush &\XSolidBrush  &\Checkmark
 & 59.9 & 24.7 & 39.1 & 30.3 & 24.5 \scriptsize{(-0.8\%)} & 31.4 \scriptsize{(-19.7\%)} & 27.5 \scriptsize{(-9.1\%)}\\
UniAD~\cite{hu2023_uniad}  & \Checkmark & \XSolidBrush &\XSolidBrush &\Checkmark
& 45.8 & 47.5 & 36.3 & 41.2 & 44.0 \scriptsize{(-7.4\%)} & 27.6 \scriptsize{(-24.0\%)} & 33.9 \scriptsize{(-17.6\%)}\\
Orion~\cite{Fu2025Orion} & \Checkmark & \XSolidBrush &\XSolidBrush  &\Checkmark
 & 77.8 & 53.0 & 52.0 & 52.5 & 51.2 \scriptsize{(-3.4\%)} & 46.0 \scriptsize{(-11.5\%)} & 48.5 \scriptsize{(-7.7\%)}\\
HiP-AD~\cite{tang2025hipadhierarchicalmultigranularityplanning} & \Checkmark &\XSolidBrush &\XSolidBrush &\Checkmark 
 & 86.8 & 74.1 & 70.7 & 72.4 & 67.1 \scriptsize{(-9.4\%)} & 56.7 \scriptsize{(-19.8\%)} & 61.5 \scriptsize{(-15.1\%)}\\
SimLingo~\cite{Renz2025cvpr} & \Checkmark & \XSolidBrush &\XSolidBrush  &\Checkmark
 & 85.1 & 82.6 & 79.3 & 80.9 & 71.7 \scriptsize{(-13.2\%)} & 55.0 \scriptsize{(-30.6\%)} & 62.2 \scriptsize{(-23.1\%)}\\
\midrule
TF++~\cite{Jaeger2023ICCV} & \Checkmark & \Checkmark &\XSolidBrush &\Checkmark
 & 84.2 & 83.3 & 78.5 & 80.8 & 75.4 \scriptsize{(-9.5\%)} & 61.1 \scriptsize{(-22.2\%)} & 67.5 \scriptsize{(-16.5\%)}\\
\midrule
\midrule
PlanT 2.0~\cite{gerstenecker2025plant2} & - & - & \Checkmark &\Checkmark
 & 92.4 & 87.8 & 85.0 & 86.4 & 73.3 \scriptsize{(-16.5\%)} & 58.0 \scriptsize{(-31.8\%)} & 64.8 \scriptsize{(-25.0\%)}\\\midrule
PDMLite-F2D & - & - & \Checkmark & \XSolidBrush & 97.0 & 95.6 & 97.0 & 96.3 & 94.0 \scriptsize{(-1.7\%)} & 95.3 \scriptsize{(-1.8\%)} & 94.6 \scriptsize{(-1.7\%)}\\

\bottomrule
\end{tabular}}\vspace{-2.0mm}
\caption{\textbf{Results on Fail2Drive.} In-Distribution evaluates on known CARLA scenarios; Generalization measures robustness under distribution shift. We include reported scores on Bench2Drive for comparison. 
}
    \vspace{-4mm}
\label{tab:main_results}
\end{table*}

\boldparagraph{Generalization Scenarios}
We introduce 17 novel scenario classes, each instantiated in multiple configurations. All scenarios introduce a distribution shift not present in standard CARLA evaluations. These shifts target different aspects of robustness, such as altered visual appearance, non-standard obstacle geometry, or high-level behavioral deviations. For structured analysis, we group them into four generalization categories, each probing a distinct failure mode. A detailed description and qualitative examples are provided in the \supp.

(1) \textbf{Robustness scenarios.}  
These scenarios introduce elements that should not influence the driving decision, for example, when a construction cone is placed in an adjacent lane or when pedestrians are positioned safely on the sidewalk. The correct behavior is to continue driving normally. These cases test whether models rely on shallow shortcut associations (e.g., “construction cone→lane change”) rather than context-aware reasoning.

(2) \textbf{Visual generalization for lateral control.}  
Here the ego vehicle must perform a lateral avoidance maneuver in unseen situations. 
Altered parked-vehicle orientations, unseen obstacle assets, and unusual object layouts test whether models trigger avoidance behavior only when confronted with familiar obstacle types.

(3) \textbf{Visual generalization for longitudinal control.}  
These scenarios test longitudinal avoidance maneuvers with an unseen causal object. 
Examples include replacing pedestrians with visually distinct animals, adding texture modifications to stop signs, or altering the appearance of leading vehicles. The goal is to test whether agents rely on genuine semantic cues rather than memorized visual templates.

(4) \textbf{Behavioral generalization scenarios.}  
These require behaviors rarely demonstrated in standard CARLA data, such as coming to a full stop and waiting when both lanes are completely blocked or maintaining a safe following distance to slow-moving pedestrians on the roadway. The ego vehicle must exhibit a high-level behavior that cannot be solved through memorized patterns alone. 

\boldparagraph{Generalization pairs}
To measure not just performance under shift but the actual generalization gap, we introduce 100 in-distribution/generalization pairs.
The in-distribution routes use unmodified CARLA scenarios, while the generalization route introduces only the targeted shift while preserving road geometry, spawn points, and traffic goals. This design isolates robustness from absolute driving performance and enables a direct computation of a generalization gap.

\boldparagraph{Metrics}
We follow the Bench2Drive evaluation protocol and report Driving Score (DS) and Success Rate (SR). Route Completion is omitted because Fail2Drive routes are intentionally short and nearly always finishable, making it a poor indicator of robustness. To quantify generalization, we compute the relative performance difference between each route pair and aggregate results across categories. Inspired by the F1-metric, we additionally report the harmonic mean of DS and SR. This harmonic mean (HM) provides a single, balanced metric for comparing models and jointly captures reductions in both Driving Score and Success Rate.

\subsection{\benchName{} Rules}
To ensure fair comparison, the following rules apply:

\begin{enumerate}
    \item \textbf{No training on \benchName{} scenarios.} Models must not use the routes, scenario definitions, or assets introduced in \benchName{} for training or fine-tuning. The benchmark serves strictly as a held-out test set.
    
    \item \textbf{External pretraining is allowed.} Pretraining on large-scale real-world datasets, internet-scale multimodal corpora, foundation models, or VLM/LLM backbones is permitted. Such general visual or linguistic knowledge is considered part of the model prior and not a violation of the benchmark.

    \item \textbf{Leaderboard entry.} We encourage users to submit final scores through the public evaluation repository via pull request. This enables consistent comparison and facilitates transparent benchmarking.
\end{enumerate}

\noindent\textbf{On pretraining leakage.} Large pretrained models may have seen visually similar objects or scene types in unrelated datasets. We do not attempt to restrict such general knowledge, as it is impractical to trace and is an important research direction. We included long-tail scenarios that are unlikely to appear in real-world driving datasets. The primary restriction is that the new benchmark scenarios themselves must not be used during training.

\begin{figure*}[t]
    \centering
    \vspace{1.0mm}
    \includegraphics[width=0.94\textwidth]{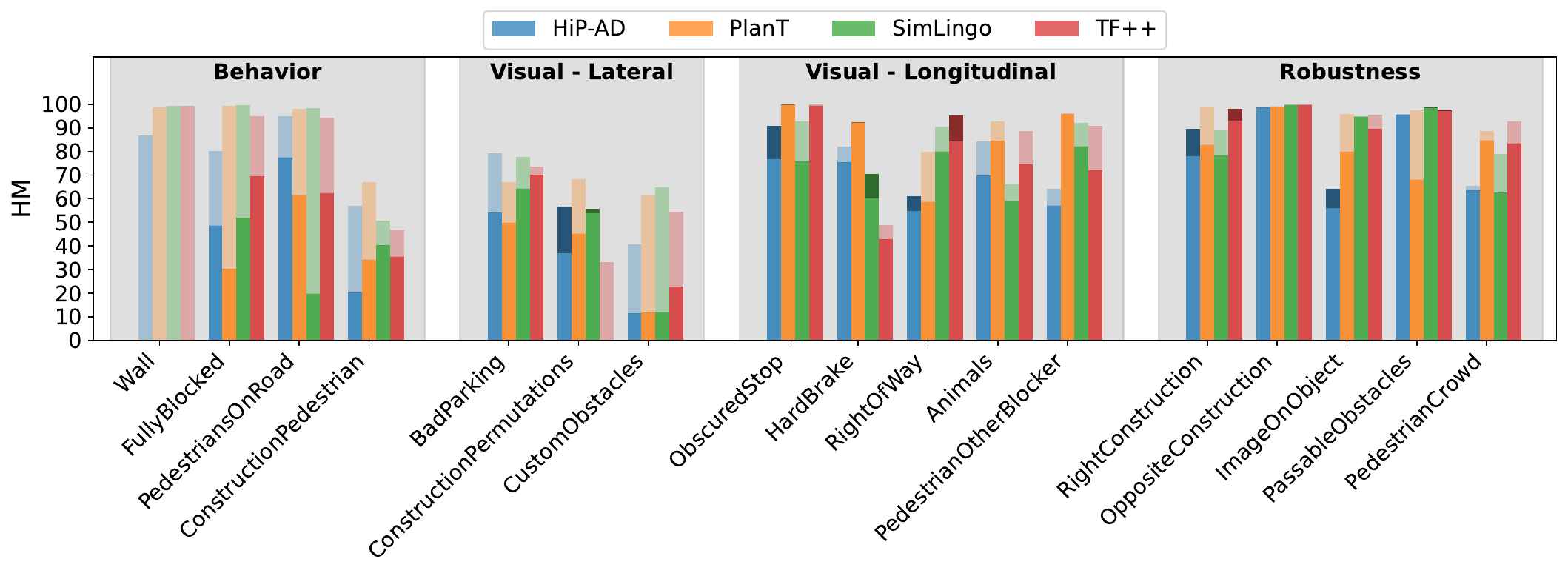}
    \vspace{-3.0mm}
    \caption{\textbf{Category-wise generalization performance.} Harmonic mean between Driving Score and Success Rate on the four scenario categories of \benchName{}. The transparent part displays the drop in performance, and the darker ones indicate an increase in score.}
    \vspace{-4mm}
    \label{fig:barplot}
\end{figure*}

\subsection{Scope and Sim2Real Interpretation}
While simulators like CARLA inherently introduce a sim-to-real gap, they remain an invaluable platform for fundamental research and principled analysis. Conducting large-scale testing of safety-critical scenarios in the real world is often infeasible, prohibitively expensive, and dangerous. Furthermore, the community is actively addressing the fidelity gap through recent advances in photorealism and domain adaptation, such as Cosmos-transfer\cite{nvidia2025cosmostransfer1conditionalworldgeneration}. Consequently, while we do not argue that results from Fail2Drive transfer directly to real-world deployment, evaluating unseen scenarios in a controlled environment remains a critical missing piece in the current landscape. Systematic stress-testing in simulation is a necessary prerequisite to quantify and compare the increasing generalization efforts by the community.
By pairing each generalization route with an in-distribution counterpart at the same location and with identical traffic configuration, we isolate the effect of a targeted structural change while keeping all other factors fixed. This design enables controlled analysis of whether policies rely on transferable driving concepts or on unreliable patterns.

\section{Analysis of State-of-the-Art Models}
\label{sec:experiments}

In this section, we investigate two questions: (i) How strongly do representative SOTA models degrade under controlled distribution shift? (ii) Which type of shift drives failures? 

We evaluate seven representative closed-loop driving models on \benchName{} to measure their robustness under controlled distribution shifts. We show how models fail to adapt to even small modifications of CARLA scenarios. The selected models span classical camera-based policies, multimodal fusion architectures, vision-language-action (VLA) systems, and privileged planners.

\begin{itemize}
  \item \textbf{TCP}~\cite{Wu2022NeurIPS} is a CNN-based baseline that drives using a single front camera, ego state, and route information.
  \item \textbf{UniAD}~\cite{hu2023_uniad} encodes six camera inputs into a BEV feature space for planning-oriented driving.
  \item \textbf{TransFuser++}~\cite{Jaeger2023ICCV} fuses LiDAR and camera data using a transformer to jointly predict driving plans and auxiliary perception tasks.
  \item \textbf{HiP-AD}~\cite{tang2025hipadhierarchicalmultigranularityplanning} predicts coarse and fine waypoints in parallel from six cameras, improving interpretability and trajectory robustness.
  \item \textbf{SimLingo}~\cite{Renz2025cvpr} is a vision-language model using only one front-facing camera and is able to output language descriptions.
  \item \textbf{Orion}~\cite{Fu2025Orion} integrates six camera views with a transformer-based fusion module, which is input to the LLM that generates multiple auxiliary tasks and an action through a generative planner.
  \item \textbf{PlanT 2.0}~\cite{gerstenecker2025plant2} uses privileged simulator information and a sparse object representation.
  \item \textbf{PDMLite-F2D} is our extension to the privileged rule-based expert PDMLite~\cite{sima2023drivelm} covering our new scenarios. See Section~\ref{sec:toolbox} for details.
\end{itemize}

For TCP and UniAD, we use the reimplementations provided by~\cite{jia2024bench}, which were trained on the Bench2Drive dataset. 
For all other models, we use the original checkpoints and evaluation code provided by the authors.
We evaluate each model using three different evaluation seeds and report averages.

\subsection{Generalization Gap}

Table~\ref{tab:main_results} reports Driving Score (DS), Success Rate (SR), and the Harmonic Mean (HM) on both in-distribution and generalization routes. We include sensor-based models (top) and privileged models (bottom). Across all seven learning-based models, we observe a consistent performance drop under the proposed shifts, with an average HM drop of 16.3\%, indicating that current CARLA-based driving stacks do not generalize reliably beyond their training distributions. 

For the remainder of the analysis, we focus on the four models that achieve reliable in-distribution performance with scores above 70: HiP-AD, SimLingo, TransFuser++, and PlanT 2.0.

The privileged learned planner PlanT 2.0 achieves the strongest in-distribution performance (86.4 HM) but suffers a 25.0\% HM drop, falling below the sensor-based model TransFuser++ on the generalization routes.

Among the sensor-based methods, SimLingo achieves the highest performance on the in-distribution routes (80.9 HM) but shows a large generalization gap of -23.1\% HM. Indicating that using VLM pre-training alone does not necessarily help for generalization. 
TransFuser++ and HiP-AD obtain a lower but still significant performance drop of 16.5\% and 15.1\%. 
TransFuser++, the only camera+LiDAR model, achieves the overall strongest performance on the generalization routes (67.5 HM).

\subsection{Failure investigation}\label{sec:failure}

To understand where the generalization failures originate, we analyze performance by scenario category (Fig.~\ref{fig:barplot}). Across categories, we observe a consistent underlying pattern: models rely on \emph{recurring CARLA-specific patterns} rather than forming general concepts of obstacles, road users, or drivable space. Once those patterns are altered, even minimally, planning performance often breaks down.

\boldparagraph{Behavior}
Behavioral generalization scenarios require models to execute previously unseen high-level actions, such as following pedestrians walking on the road, stopping and waiting for completely blocked roads, or slowing for crossing pedestrians while navigating around a construction site.

Across all four models, we observe the largest generalization gap of any category, with an average drop of -53.6\% HM. Despite strong in-distribution performance, all models struggle to deviate from memorized lane-following behavior when confronted with novel situations.

\begin{figure}[t]
    \centering
    \includegraphics[trim={15cm 18cm 15cm 20cm}, clip, height=2.9cm]{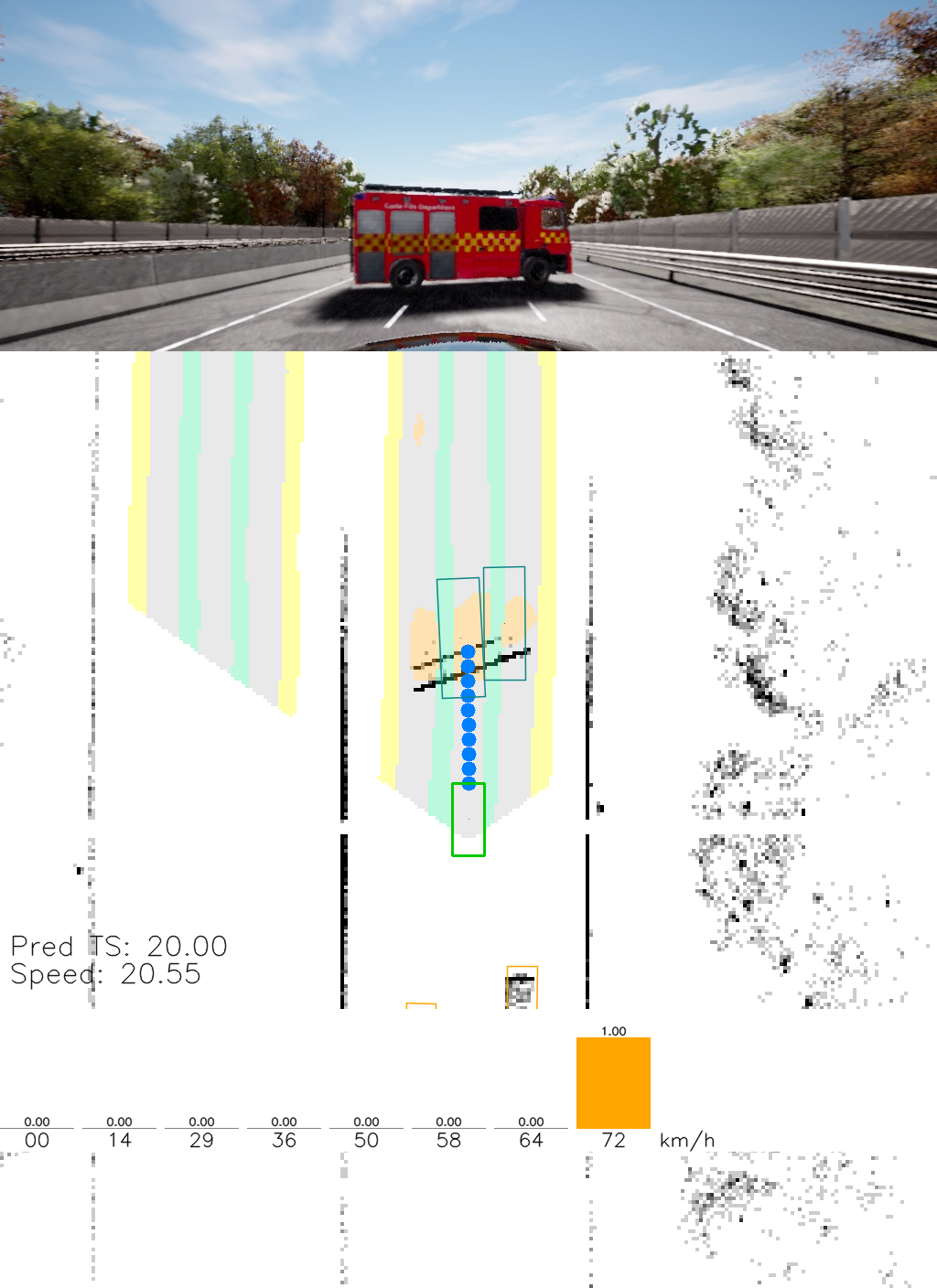}
    \hfill
    \includegraphics[height=2.6cm]{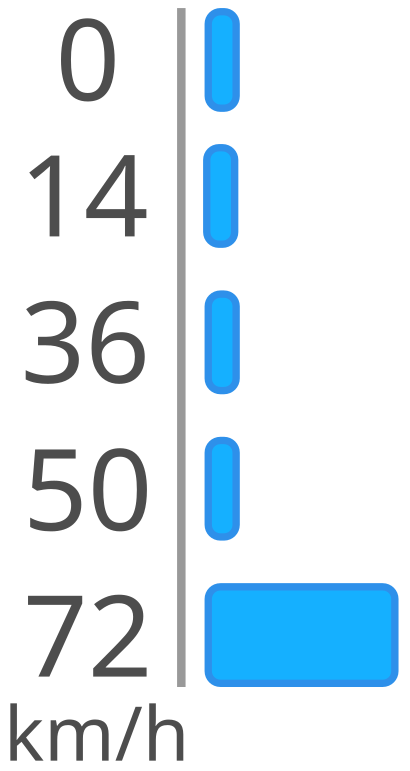}
    \hfill
    \includegraphics[trim={12.2cm 36.2cm 12.7cm 4.5cm}, clip, height=2.9cm]{qualitative/tfpp_fullyblocked.png}%
    \caption{\textbf{Blocked road.} TransFuser++ correctly perceives a vehicle, but predicts full-speed driving, causing a collision. 
    }\vspace{-2mm}
    \label{fig:perception_failures}
\end{figure}

\textit{\texttt{FullyBlocked}: failure to stop for a clearly obstructed road.} In \texttt{FullyBlocked}, the road is entirely obstructed by trucks, vans, or hay bales. Surprisingly, even models with access to privileged information or rich sensor inputs fail: PlanT, despite receiving ground-truth object positions, drops by -69.5\% HM, the largest degradation of all models, showing heavy overfitting to the exact object sizes and locations. TransFuser++, the only LiDAR-based model, shows the smallest decline (-26.7\% HM) yet still fails to stop reliably (Fig. \ref{fig:perception_failures}).

\begin{figure}
    \centering
    \includegraphics[width=0.8\linewidth]{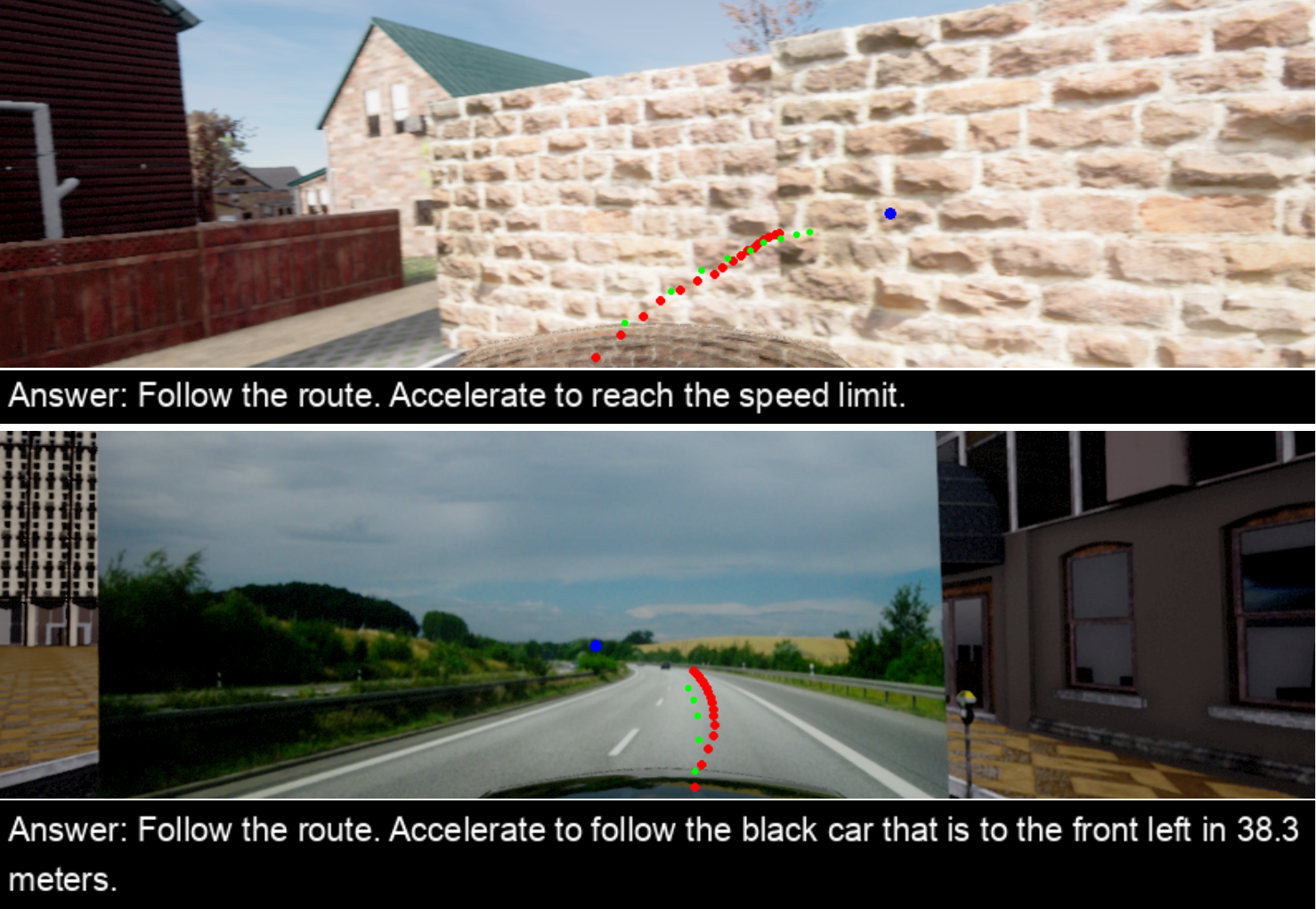}%
    \caption{\textbf{Wall collisions.} SimLingo is reacting to a wall blocking the road. (Top) SimLingo tries to avoid the obstacle by changing lanes, causing a collision. (Bottom) SimLingo does not recognize the picture wall and instead adjusts its predictions for the road depicted on the wall, even correctly identifying a vehicle in the image.}\vspace{-2mm}
    \label{fig:simlingo_wall}
\end{figure}

\textit{Wall: adversarial appearance changes break all models.} The \texttt{Wall} scenario is an extreme version of the previous scenario, featuring a large wall blocking the road. In 4 out of 5 cases, it displays a full-size printed image of a road. All models collapse from around 100 HM to 0.0 HM.
Multiple policies misinterpret the printed road in the image as continuation of the real drivable space (Fig.~\ref{fig:simlingo_wall}).
Most notably, TransFuser++, equipped with LiDAR, also fails to detect or react to the obstacle, again demonstrating the reliance on known cues. 

\begin{figure}
    \centering
    \includegraphics[trim={5cm 18cm 5cm 6cm}, clip, width=0.8\linewidth]{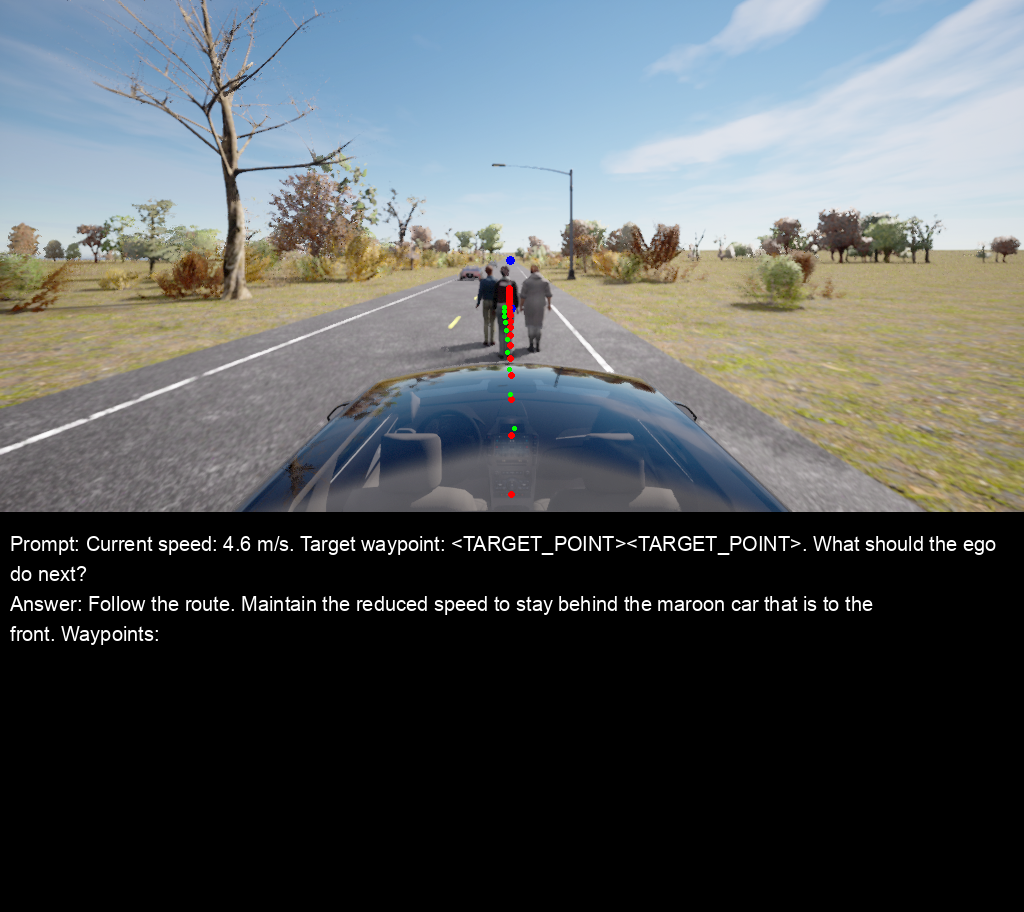}
    \includegraphics[trim={0 9cm 5cm 20.5cm}, clip, width=0.8\linewidth]{qualitative/simlingo_pedestrians.png}%
    \caption{\textbf{Pedestrian collision.} SimLingo predicts a reduced velocity but fails to correctly identify the pedestrians on the road, eventually causing a collision.}\vspace{-2mm}
    \label{fig:simlingo_pedonroad}
\end{figure}

\textit{PedestriansOnRoad: inability to stay behind slow agent.}
In \texttt{PedestriansOnRoad}, a small group of pedestrians walks along the ego lane. Models are expected to slow down and follow at a safe distance, or overtake safely. However, models frequently approach too closely and collide when pedestrians pause.
HiP-AD sometimes attempts to overtake the pedestrians with varying degree of success.
SimLingo performs particularly poorly, dropping from 98.50 to 19.68 HM with collisions in 87\% of cases. Its language-action module often hallucinates a nonexistent car or cyclist to follow (Fig.~\ref{fig:simlingo_pedonroad}), showing overfitting to the language used during training.
The consistent failure across models shows that vehicle-following cues do not generalize to non-vehicle agents.

Across all behavioral scenarios, models display a common pattern:
\textbf{When encountering unseen or ambiguous situations, models default to memorized driving patterns, typically lane-following}, rather than adopting fallback strategies such as stopping or waiting. This tight coupling between learned behaviors and CARLA-specific scenario templates highlights a fundamental limitation of current driving systems: even strong models do not yet possess a generalizable notion of high-level driving behavior.

\begin{figure}[t]
    \centering
    \begin{subfigure}{0.39\linewidth}
    \centering
    \includegraphics[trim={12cm 17cm 12cm 6cm}, clip, height=2.2cm]{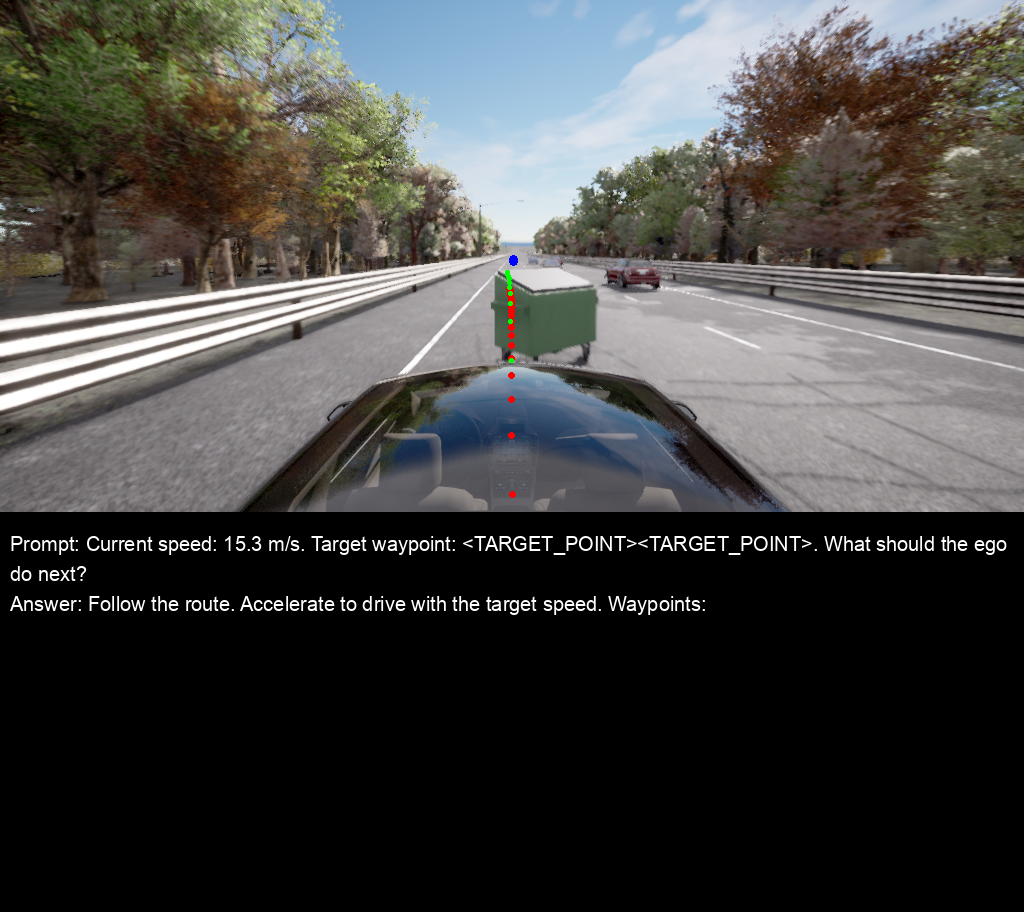}
    \caption{SimLingo}
    \end{subfigure}
    \hfill
    \begin{subfigure}{0.59\linewidth}
    \centering
    
    \includegraphics[trim={15.5cm 18cm 16.5cm 23.5cm}, clip, height=2.2cm]{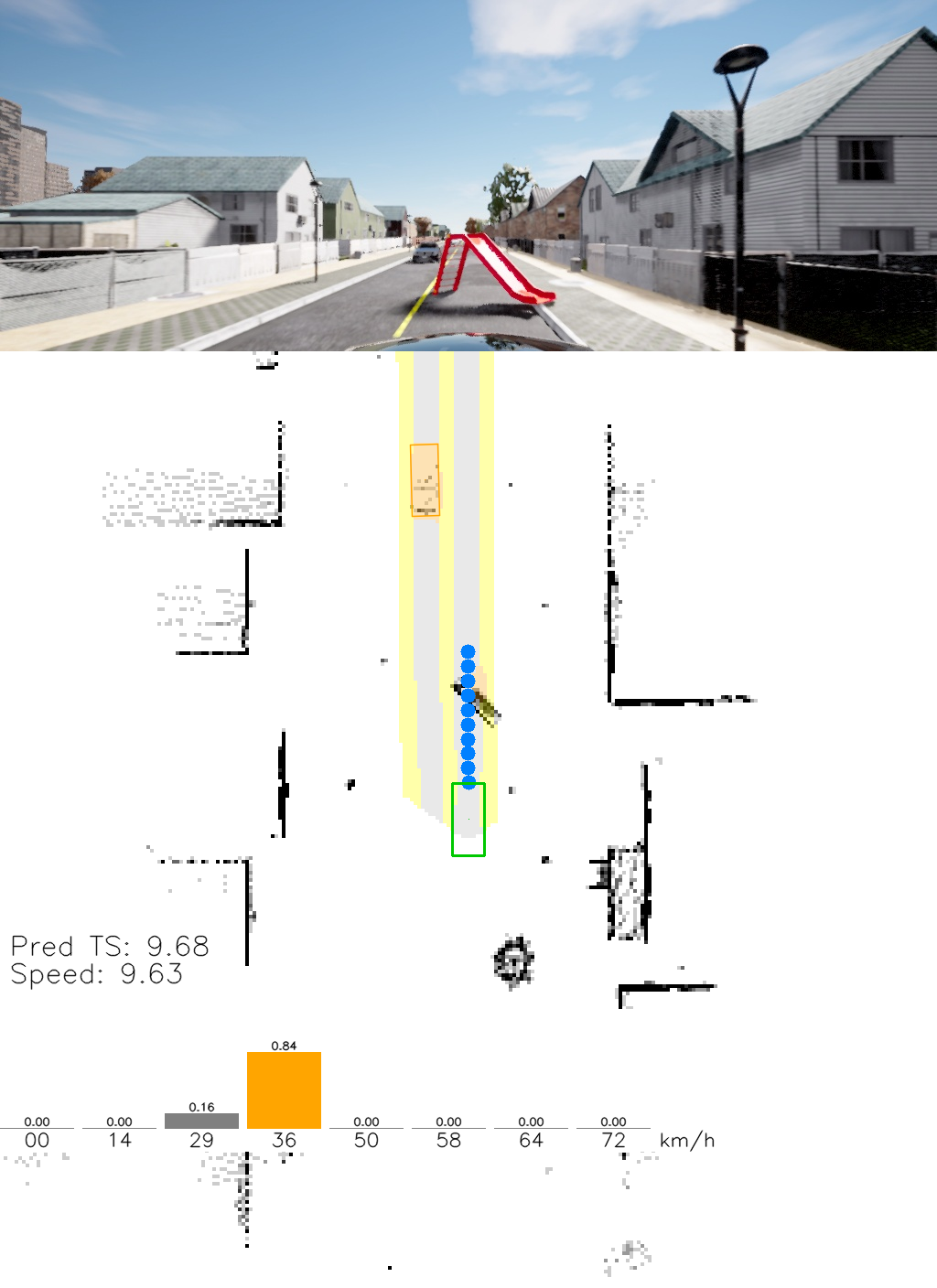}
    \includegraphics[trim={52cm 17cm 48cm 23cm}, clip, height=2.2cm]{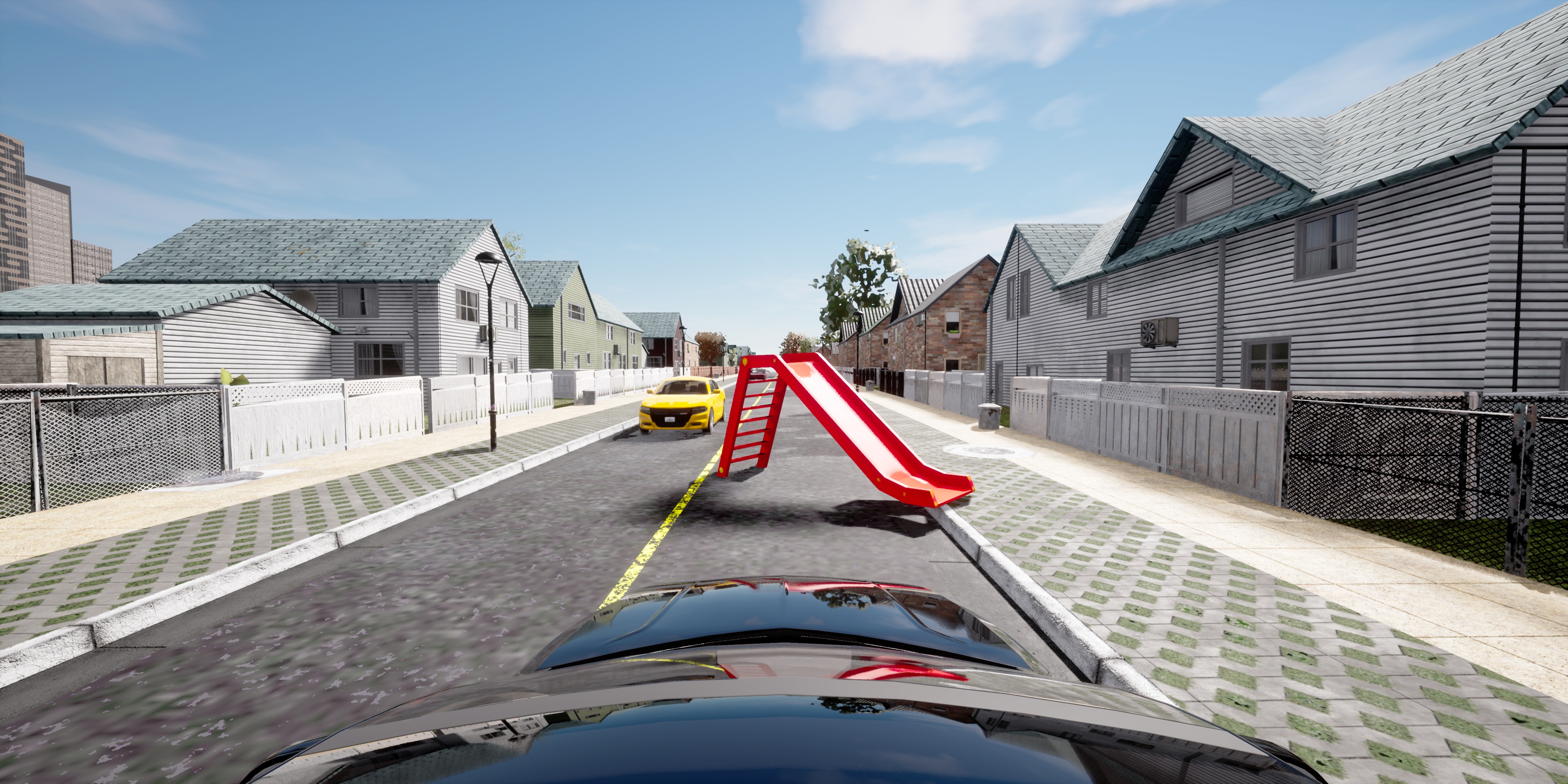}
    \caption{TransFuser++}
    \end{subfigure}%
    \caption{
\textbf{Obstacle failures.} SimLingo and TransFuser++ fail to detect and avoid clearly visible obstacles. TransFuser++'s LiDAR visualisation shows the apparent obstacle in the LiDAR data and no bounding box prediction being made.}\vspace{-2mm}
    \label{fig:obstacle_failures}
\end{figure}

\boldparagraph{Visual - Lateral} 
These scenarios evaluate whether models can identify and navigate around obstacles whose appearance, geometry, or spatial layout differs from CARLA defaults. These scenarios require an inherently complex behavior: in-distribution scores are already low across all models, and introducing even small appearance shifts leads to substantial additional degradation of on average -33.58\% HM.

\textit{CustomObstacles: failure to avoid unseen obstacles.}
The \texttt{CustomObstacles} scenario is the most challenging across all lateral tasks.
Even when obstacles are large and unambiguously visible in both RGB and LiDAR (Fig.~\ref{fig:obstacle_failures}), all models fail to avoid them robustly. Performance drops to 11.44 HM (HiP-AD), 11.73 HM (PlanT), 11.90 HM (SimLingo), and 22.78 HM (TF++).
This failure illustrates that avoidance behaviors are not triggered by the spatial presence of obstacles in the drivable lane. Instead, models depend on familiar CARLA-specific obstacle templates (cones, traffic warnings, standard vehicle meshes). Importantly, many of the new scenarios use assets from the same source as the original CARLA assets, suggesting that the gap is primarily structural rather than just a distribution shift in texture.

\begin{figure}[t]
    \centering
    \begin{subfigure}{0.7\linewidth}
    \includegraphics[trim={15.5cm 18cm 16.5cm 23.5cm}, clip, height=2.6cm]{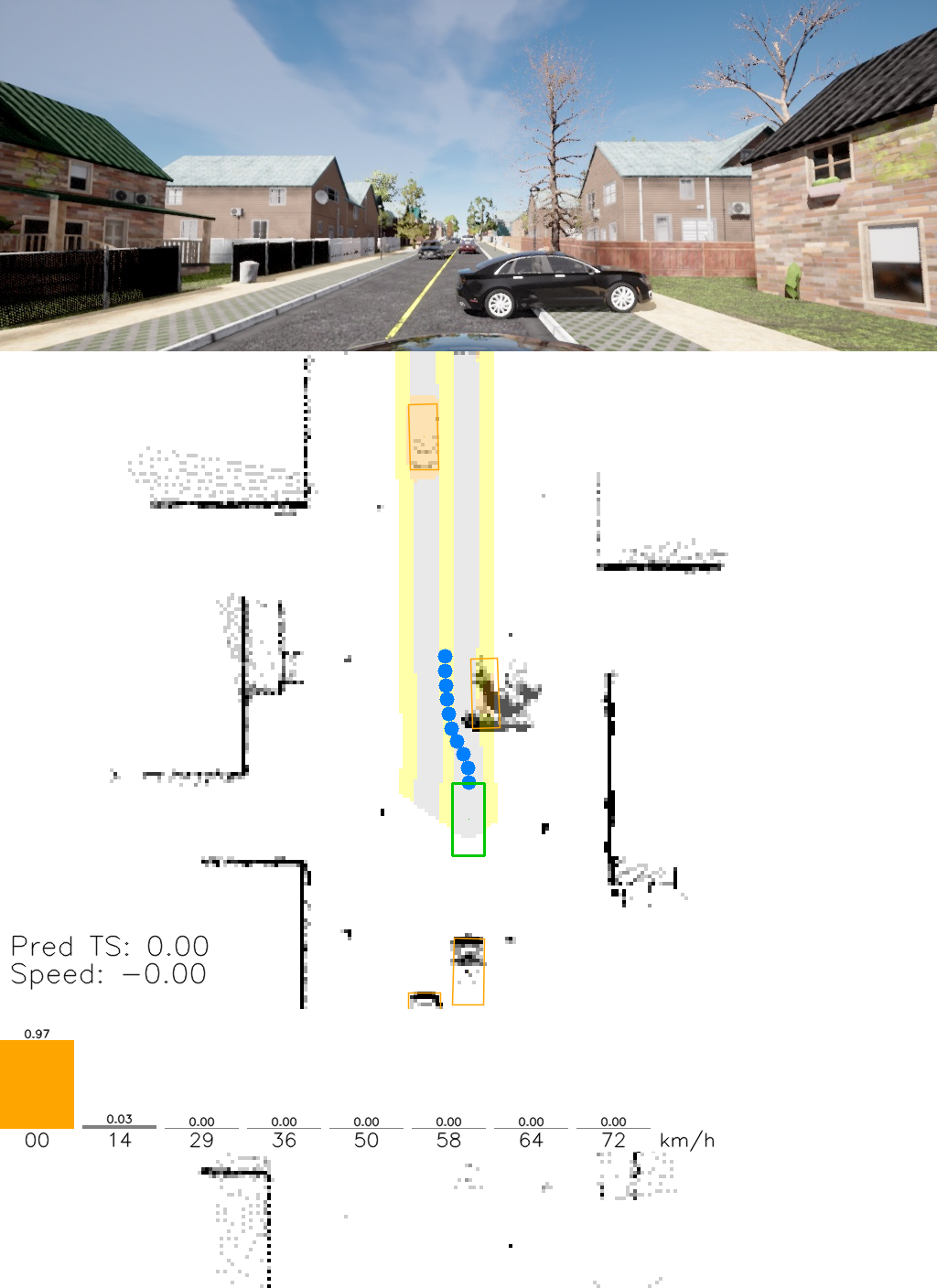}
    \includegraphics[trim={15cm 37cm 12cm 6.3cm}, clip, height=2.6cm]{qualitative/tfpp_bbox2.png}
    \caption{TransFuser++}
    \end{subfigure}
    \hfill
    \begin{subfigure}{0.26\linewidth}
    \includegraphics[trim={59cm 8.5cm 6cm 5cm}, clip, height=2.6cm]{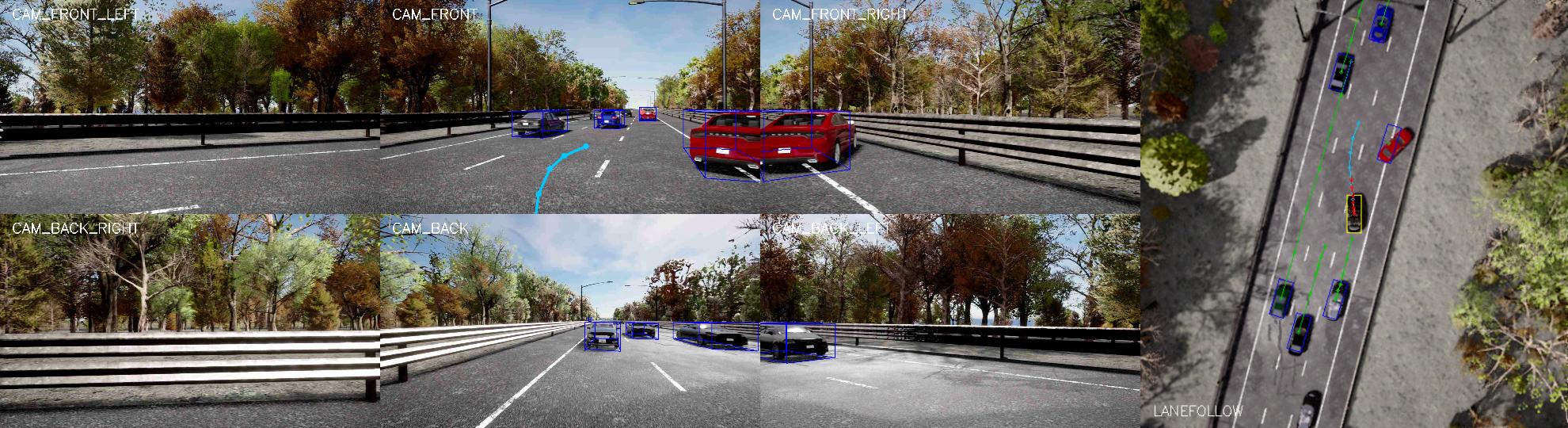}
    \caption{HiP-AD}
    \end{subfigure}%

    \caption{\textbf{Perception failures.} TransFuser++ and HiP-AD fail to correctly perceive the parked vehicle's orientation, defaulting to the orientation seen in CARLA demonstrations, but are thereby able to solve the scenario.}\vspace{-2mm}
    \label{fig:perception_overfit}
\end{figure}

\textit{BadParking: orientation priors in perception.}
In \texttt{BadParking}, the same vehicle assets are placed in unusual orientations. While this requires only small deviations in the planned trajectory, performance still drops noticeably.
Models with explicit perception heads (HiP-AD and TransFuser++) often predict the default CARLA parked-car orientation, regardless of the actual rotation (Fig.~\ref{fig:perception_overfit}).
Although TF++ sometimes still manages to avoid the obstacle despite the incorrect perception orientation, the systematic misalignment reveals a strong geometric prior extending beyond planning into perception. 
These results show that perception is not just sensitive to appearance, but also to geometric diversity.

\textit{ConstructionPermutations: removal of cues breaks avoidance.}
The \texttt{ConstructionPermutations} scenarios expose an even deeper reliance on CARLA-specific symbolic cues.
Small modifications to the construction layout, like removing the warning sign or some of the cones, or replacing assets with visually similar variants, cause dramatic failures.
TransFuser++ collapses from 33.15 HM to 0.00 HM when the main warning sign is removed.
Since SimLingo only drops from 75.74 to 56.12 HM (-25.9\%), the TF++ result indicates overfitting to the lidar signature of the CARLA construction scenario.
PlanT instead relies on the traffic cones placed at the side of constructions. Removing the cones, leaving only the big construction warning, causes scores to drop from 100.00 to 0.00 HM, despite having ground-truth object positions.

These large drops indicate that models \textbf{do not reason about drivable space directly}. Instead, they \textbf{rely on hard-coded pattern associations}, such as “a construction warning sign means: do a lane change”.
When these patterns break down, avoidance behavior simply does not trigger, even when the obstacle itself remains clearly visible.

\begin{figure}[t]
    \centering
    \includegraphics[trim={20cm 10.5cm 37cm 4.5cm}, clip, width=0.7\linewidth]{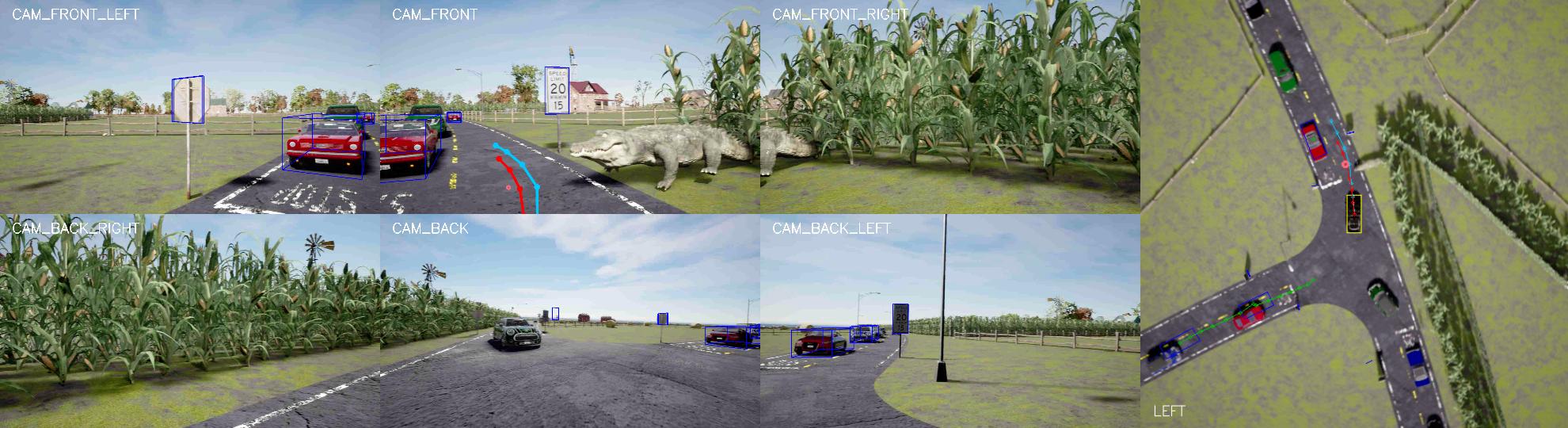}
    \includegraphics[trim={20cm 11.5cm 37cm 3.5cm}, clip, width=0.7\linewidth]{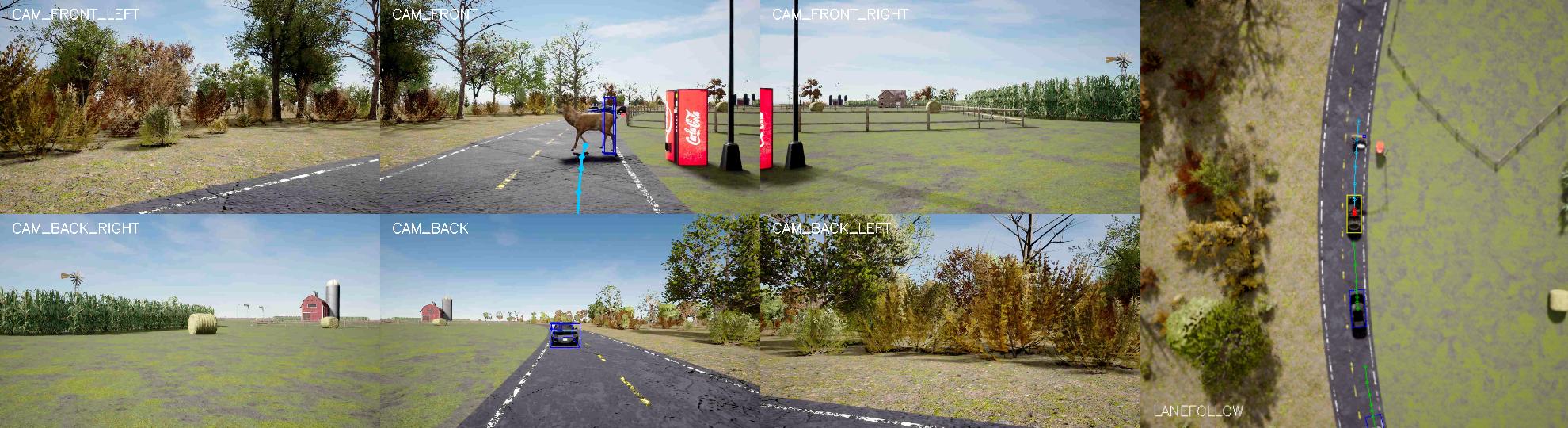}%
    \caption{\textbf{Animal perception failure.} HiP-AD is more likely to detect an animal as a pedestrian when it has clearly visible, human-like legs, positioning the bounding box at its legs. %
    }\vspace{-2mm}
    \label{fig:pedestrian_legs}
\end{figure}

\boldparagraph{Visual - Longitudinal}
These scenarios evaluate whether models can slow down or stop when the causal object (pedestrian, leading vehicle, or traffic sign) undergoes appearance or geometric changes that were not present during training. 

All four models maintain good performance, with a moderate average HM drop of 7.2\%. This indicates that existing policies can transfer several visual cues of “something ahead requires slowing down”, a promising sign of within-CARLA generalization beyond a narrow set of CARLA assets.

\textit{Animals: behavior gets brittle with visual and geometric deviation.}
When replacing pedestrians with animals, the object detections from TF++ and HipAD are more likely to trigger correct behavior for animals with human-like structure (e.g., deer, zebras). Compact or non-upright shapes  (e.g., pigs, crocodiles) are frequently misclassified or ignored (Fig.~\ref{fig:pedestrian_legs}).

\textit{RightOfWay: implicit behavioral priors.}
In \texttt{RightOfWay}, where emergency vehicles take the ego vehicles' right of way, replacing the emergency vehicle with a regular car noticeably increases collisions for SimLingo and PlanT. Both appear to assume that only emergency vehicles violate right-of-way, exposing a behavioral shortcut rather than a robust interpretation of motion cues.

\textit{ObscuredStop: distribution shift in sign detection.} In \texttt{ObscuredStop}, textures are placed on stop signs to evaluate stopping behavior under shift. Three out of four models are unaffected by these visual changes, with only SimLingo exhibiting an increase in stop infractions from 6\% to 33\% of scenarios. The scenarios where these failures occur include the snow and sticker textures (Fig.~\ref{fig:teaser}), which have the highest amount of occlusion. %
The introduced variations do not cause SimLingo to fail to identify the stop signs entirely; instead, they cause it to stop further away from the stopping line, leading CARLA to issue a penalty. 

\textit{Success cases and their implications}
While longitudinal scenarios show clear weaknesses, they also highlight where the models excel.
HiP-AD and TransFuser++ remain stable when modifying visual appearance without altering geometry (e.g., \texttt{ObscuredStop}), suggesting robustness to small texture noise.
Models also perform well when the causal object retains a similar scale and pose (e.g. some of the animals).

\boldparagraph{Robustness}
Robustness scenarios test a model's ability to ignore irrelevant environmental influences. Examples include construction assets positioned off the drivable lane, static pedestrian crowds standing on sidewalks, or printed imagery placed in non-actionable locations. Since these elements do not require behavioral adjustment, the ideal policy is to continue driving normally, without unnecessary deceleration, lane changes, or hesitation.

Across all models, robustness scenarios exhibit the strongest generalization performance of any category. Avoiding overreaction appears substantially easier than generating new behaviors, with most models maintaining high HM scores and only minor degradation under environmental shifts. 

PlanT and SimLingo show the largest behavioral deviations in this category. While overall scores remain high, both models exhibit an average velocity drop of approximately 10\% under generalization scenarios, whereas the other methods maintain stable speeds. This indicates heightened policy uncertainty and sensitivity to irrelevant objects.

\begin{figure}[t]
    \centering
    \includegraphics[trim={3cm 1.5cm 3cm 3cm}, clip, height=2.8cm]{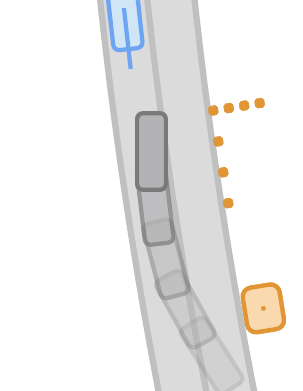}
    \hspace{1cm}
    \includegraphics[trim={3cm 2cm 3cm 2.5cm}, clip, height=2.8cm]{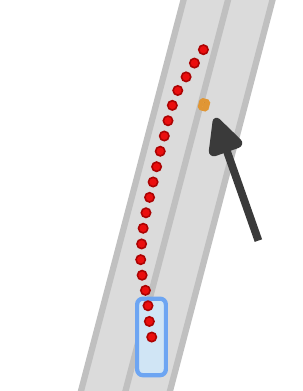}
    \caption{\textbf{PlanT~2.0 robustness.} PlanT performs an unnecessary avoidance maneuver for a construction (left) and a mailbox that both do not block its path, risking vehicle collisions. \label{fig:plant_robustness}}\vspace{-2mm}
\end{figure}

In the \texttt{RightConstruction} scenario, a construction site is located in an adjacent lane. Although the obstacle does not obstruct the ego trajectory, several models prepare for a lane change or reduce speed before resuming normal driving. SimLingo reacts strongly, slowing down to prepare to cross into the left-adjacent lane but eventually continues driving in most cases, whereas PlanT~2.0 executes a full evasive maneuver, as if the obstacle were directly in its path (Fig.~\ref{fig:plant_robustness}).

A similar pattern emerges in \texttt{PassableObstacle} scenarios, where obstacles appear on or near the road without intersecting the ego trajectory. For instance, when a single mailbox is placed on the centerline (Fig.~\ref{fig:plant_robustness}), PlanT~2.0 performs an extensive evasive maneuver, as the placement of this small object is similar to the traffic cones in a CARLA construction scenario. This sensitivity of PlanT~2.0 to construction cones has already been discovered in open-loop evaluations by the PlanT~2.0 authors. Our closed-loop experiments confirm this behavior.

These results underscore that \textbf{some models fail to generalize to distractions in the scenes}, with PlanT~2.0 being particularly sensitive.

\section{\benchName{} Toolbox} \label{sec:toolbox}

Beyond evaluation, \benchName{} provides a toolbox that enables controlled scenario extension, new benchmarks, and construction of a diverse large-scale dataset.

\boldparagraph{Custom simulator}
\benchName{} includes a modified build of CARLA 0.9.15 that adds a set of new assets intended to support controlled distribution shifts. The release includes 17 animal assets with correct animations, which allows evaluation of whether policies generalize the concept of “yielding to vulnerable road users” beyond the specific pedestrian meshes commonly used in CARLA.

We also include three families of visual obstacle assets. The first consists of four large-scale ``image walls'' that appear as physical obstacles blocking the road. Three walls display high-resolution driving-scene photographs, while one displays a brick texture. The second family introduces five variants of stop signs with structural or texture faults (e.g., altered or partially missing graphics). The third set provides two images of a running child at different scales and one image of a red traffic light, which can be placed as flat surfaces in the scene to test whether a model treats printed imagery as actionable scene content. Finally, the STOP road marking in Tile (3,2) of Town13 is removed, enabling evaluation of whether a policy relies on surface markings rather than sign geometry.

\boldparagraph{Customizable scenarios}
\benchName{} parameterizes several existing CARLA scenarios so that users can reconfigure them without editing the scenario code. 
This gives users broad creative freedom in scenario authoring and enables controlled \emph{counterfactual scenarios} by varying one factor at a time while keeping route context fixed.

\boldparagraph{Expert policy (PDMLite-F2D)}
We extend the rule-based expert PDMLite~\cite{sima2023drivelm} to solve \benchName{} scenarios. 
Users can apply PDMLite-F2D as a \emph{solvability check} to validate newly designed scenarios before benchmarking learning-based models.
It also enables collection of high-quality privileged demonstrations
and serves as a reference policy when debugging scenario logic. 

\section{Conclusion and Limitations}
\label{sec:conclusion}

We introduced Fail2Drive, a benchmark for evaluating generalization in CARLA. Our benchmark includes novel scenarios, unseen assets, and customization tools designed to foster future research on robust autonomous driving. Through an extensive analysis of seven recent driving models, we reveal widespread overfitting and shortcut learning, uncover unexpected failure modes, and highlight key directions for advancing generalization in end-to-end driving systems. 

In addition, our findings expose systematic gaps in current evaluation practices, which rarely probe robustness under distribution shifts. We hope that Fail2Drive raises awareness of these deficiencies and encourages the community to adopt more comprehensive OOD stress tests as part of the standard evaluation for autonomous driving models.

\boldparagraph{Limitations} We acknowledge several limitations of our work. The CARLA simulator provides only pseudo-realistic simulations, leaving uncertainties about the transferability of our results to the real world. Our claims are therefore limited to controlled closed-loop analysis in simulation: robustness in CARLA is not sufficient for real-world robustness, but we argue it is a necessary prerequisite. The paired in-distribution/generalization design also means that we primarily evaluate relative robustness under controlled structural shifts independent of absolute realism. While we provide rare, unseen scenarios, the problem of long-tail scenarios can, by definition, never be fully resolved through testing. 

\boldparagraph{Acknowledgements}
This project was supported by the DFG EXC number 2064/1 - project number 390727645 and by the German Federal Ministry for Economic Affairs and Climate Action within the project "NXT GEN AI METHODS - Generative Methoden f\"ur Perzeption, Pr\"adiktion und Planung". We thank the International Max Planck Research School for Intelligent Systems (IMPRS-IS) for supporting K. Renz.

{
    \small
    \bibliographystyle{ieeenat_fullname}
    \bibliography{bib/strings,bib/bibliography_org,bib/final_bibliography_dblp,bib/bibliography_long}
}

\clearpage
\setcounter{page}{1}
\appendix

\newgeometry{
  top=1.5in,
  bottom=1.25in,
  left=1.5in,
  right=1.5in
}
\onecolumn

{\centering
\Large
\textbf{\thetitle}\\
\vspace{0.5em}Supplementary Material \\
\vspace{1.0em}}

\section{Scenario description}
\label{sec:scenarios}
We show one example of the in-distribution/ generalization pairs of each new scenario class, together with a detailed description. The top image is always the in-distribution example, and the bottom is a new generalization sample.

\begin{enumerate}[label=\textbf{\arabic*.}]
    \item \textbf{BadParking}\\[1mm]
    \begin{minipage}{0.3\textwidth}
    \includegraphics[width=0.98\textwidth]{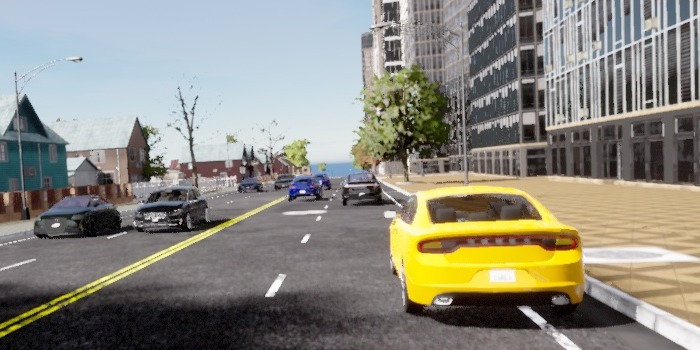}
    \includegraphics[width=0.98\textwidth]{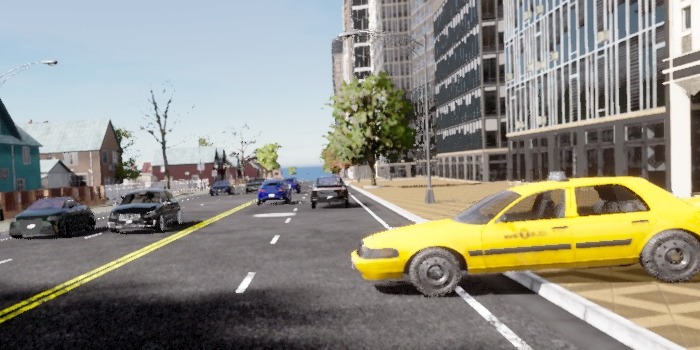}
    \end{minipage}
    \begin{minipage}{0.7\textwidth}
    A parked vehicle partially occludes the ego lane. Unlike the standard CARLA parked-vehicle scenario, which always places the vehicle in the same position, our variant can be defined with any orientation, location and asset. This is meant to challenge models' spatial understanding with known obstacles. The standard \textit{ParkedObstacle} scenario serves as an in-distribution sample.
    \end{minipage}\\

    \item \textbf{ConstructionPermutations}\\[1mm]
    \begin{minipage}{0.3\textwidth}
    \includegraphics[width=0.98\textwidth]{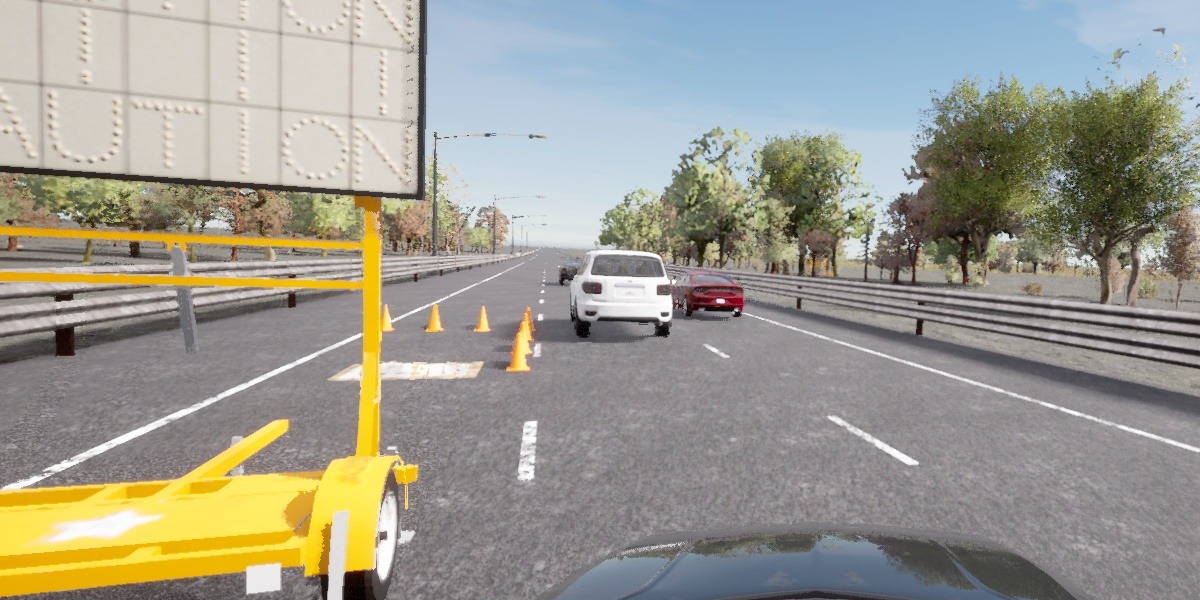}
    \includegraphics[width=0.98\textwidth]{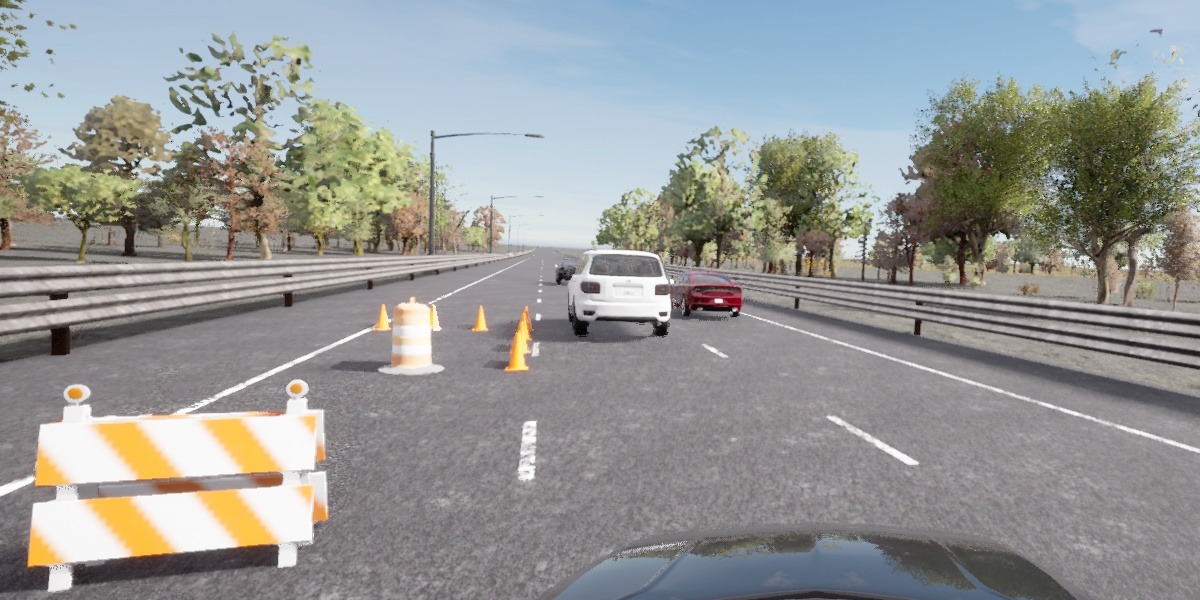}
    \end{minipage}
    \begin{minipage}{0.7\textwidth}
    A modified version of the standard \textit{ConstructionObstacle}, where construction assets can be replaced or removed, isolating dependencies on specific parts of construction sites. The in-distribution sample is defined by the default \textit{ConstructionObstacle}.
    \end{minipage}\\

    \item \textbf{CustomObstacle}\\[1mm]
    \begin{minipage}{0.3\textwidth}
    \includegraphics[width=0.98\textwidth]{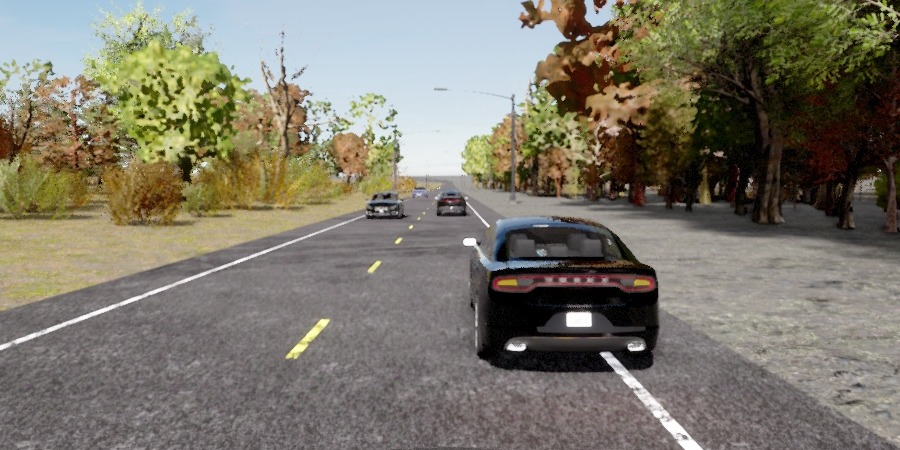}
    \includegraphics[width=0.98\textwidth]{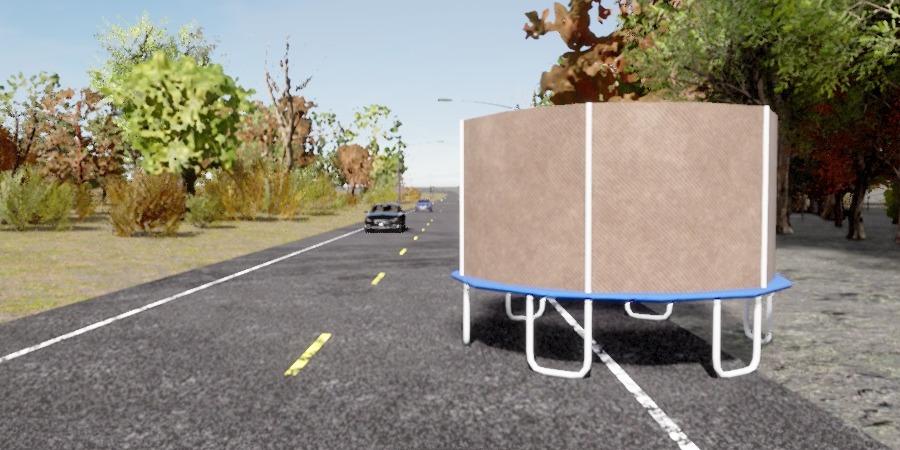}
    \end{minipage}
    \begin{minipage}{0.7\textwidth}
    Fully customizable obstacles block the road. The obstacles can be defined by any number of CARLA assets at arbitrary locations and orientations, enabling testing of generalization to unseen objects and structures. Depending on the obstacle size, a \textit{ParkedObstacle} or \textit{ConstructionObstacle} is used as an in-distribution sample.
    \end{minipage}\\

\newpage
    \item \textbf{ObscuredStop}\\
    \begin{minipage}{0.3\textwidth}
    \includegraphics[width=0.98\textwidth]{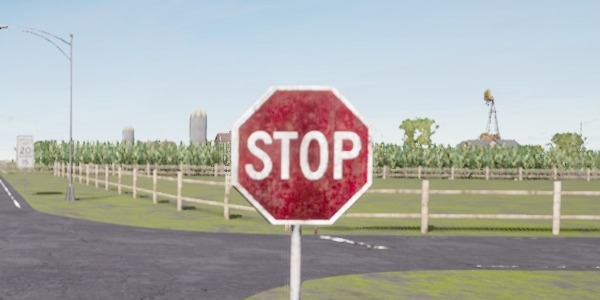}
    \includegraphics[width=0.98\textwidth]{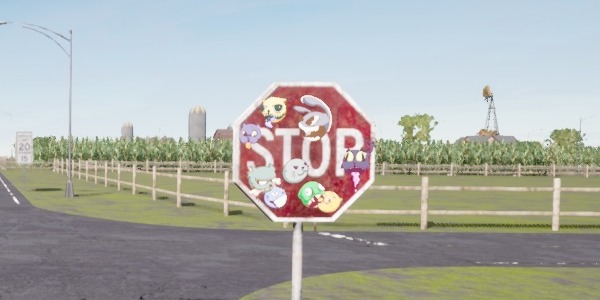}
    \end{minipage}
    \begin{minipage}{0.7\textwidth}
    Occlusions are placed on stop signs when entering an intersection, challenging the visual traffic sign detection. 5 different occlusions are included with Fail2Drive and any CARLA asset can be used. The in-distribution sample is defined by including the scenario with no occlusion.
    \end{minipage}\\

    \item \textbf{HardBrakeNoLights}\\
    \begin{minipage}{0.3\textwidth}
    \includegraphics[width=0.98\textwidth]{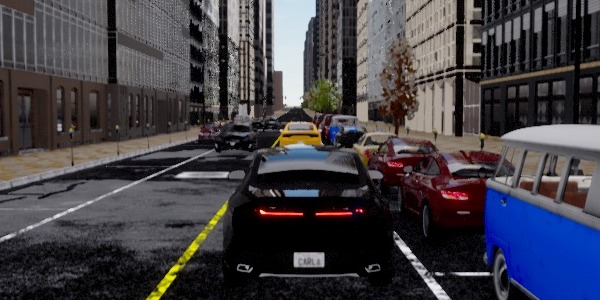}
    \includegraphics[width=0.98\textwidth]{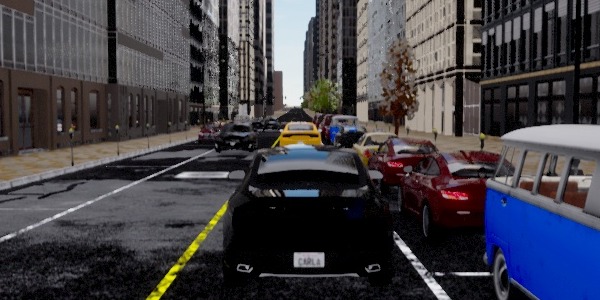}
    \end{minipage}
    \begin{minipage}{0.7\textwidth}
    The leading vehicle suddenly brakes with disabled brake lights, testing if models can judge distance and deceleration without relying on this cue. The classic \textit{HardBrake} scenario with active brake lights is used as the in-distribution sample.
    \end{minipage}\\

    \item \textbf{RightOfWay}\\
    \begin{minipage}{0.3\textwidth}
    \includegraphics[width=0.98\textwidth]{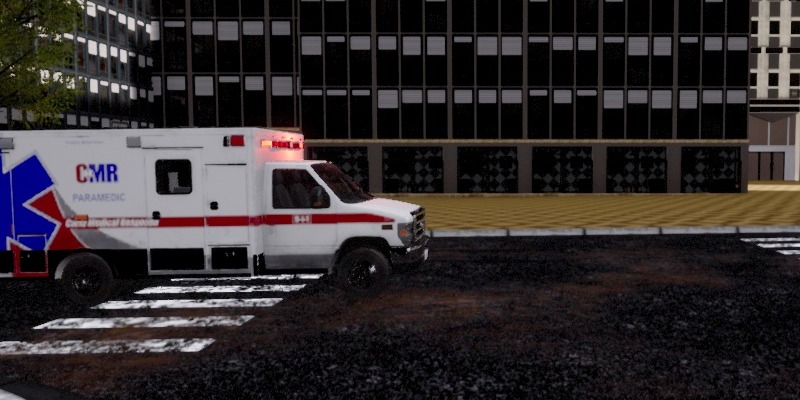}
    \includegraphics[width=0.98\textwidth]{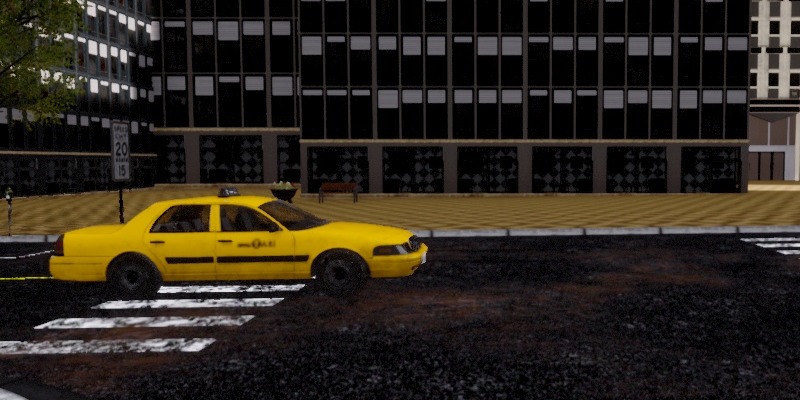}
    \end{minipage}
    \begin{minipage}{0.7\textwidth}
    A custom vehicle takes the ego vehicle's priority while crossing a junction. Since CARLA includes this scenario only with emergency vehicles, our variations test whether models only yield to emergency vehicles or generalize to other traffic participants. The emergency vehicle scenarios serve as the in-distribution sample. 
    \end{minipage}\\

    \item \textbf{Animals}\\
    \begin{minipage}{0.3\textwidth}
    \includegraphics[width=0.98\textwidth]{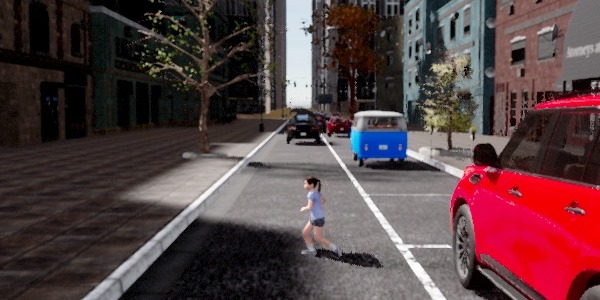}
    \includegraphics[width=0.98\textwidth]{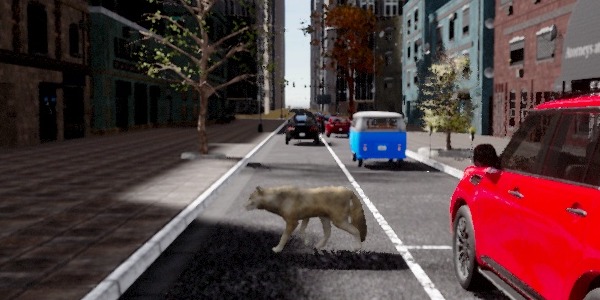}
    \end{minipage}
    \begin{minipage}{0.7\textwidth}
    An animal crosses the road, forcing the ego vehicle to react, testing if models are able to generalize to actors with other appearances and shapes as pedestrians. Fail2Drive introduces 17 animal assets that can be used for all pedestrian scenarios. By default, CARLA includes only pedestrians, which are used for the in-distribution scenario.
    \end{minipage}\\

\newpage

    \item \textbf{PedestrianOtherBlocker}\\
    \begin{minipage}{0.3\textwidth}
    \includegraphics[width=0.98\textwidth]{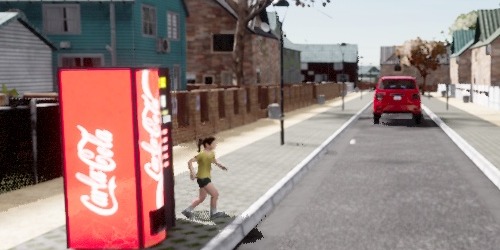}
    \includegraphics[width=0.98\textwidth]{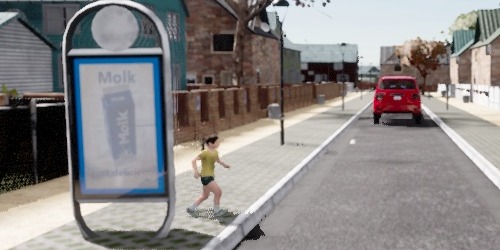}
    \end{minipage}
    \begin{minipage}{0.7\textwidth}
    A pedestrian emerges from behind an unseen object to cross the road, evaluating whether models overfit to expect pedestrians only from certain objects. The in-distribution scenario uses the default CARLA assets.
    \end{minipage}\\

    \item \textbf{RightConstruction}\\
    \begin{minipage}{0.3\textwidth}
    \includegraphics[width=0.98\textwidth]{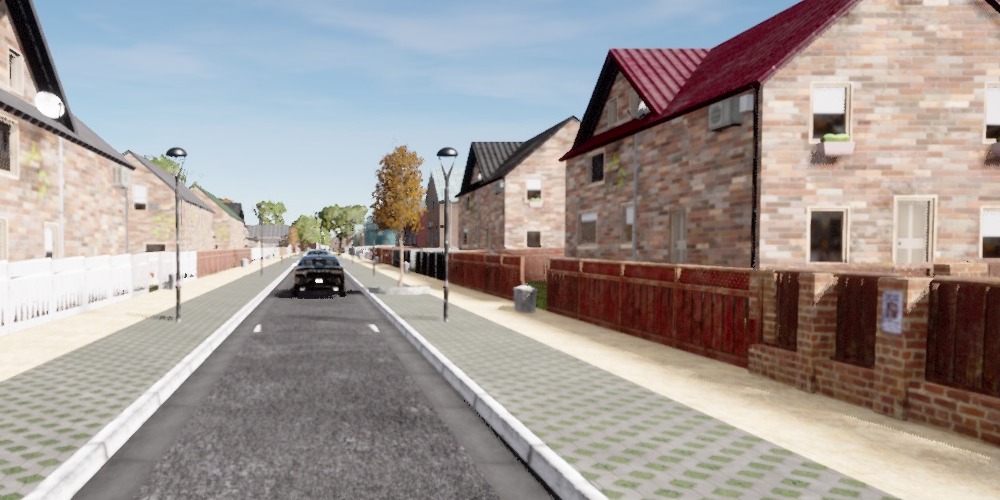}
    \includegraphics[width=0.98\textwidth]{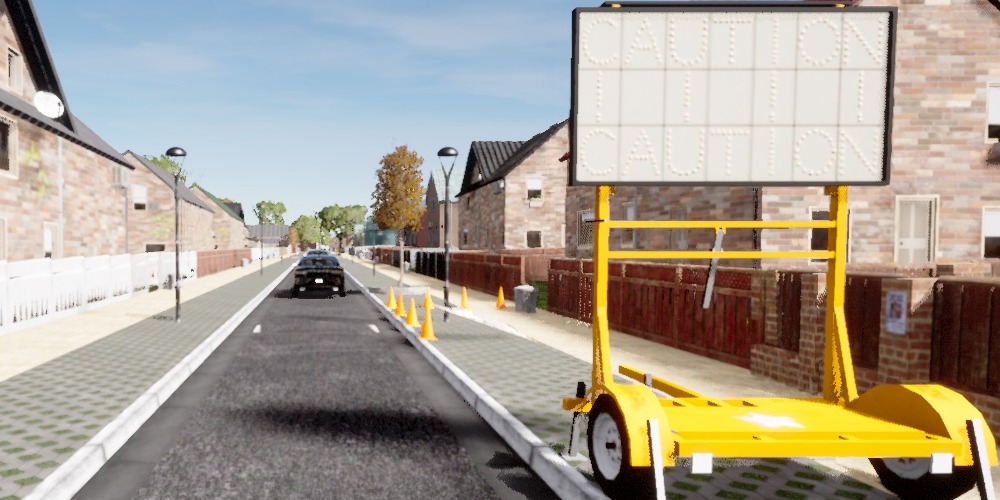}
    \end{minipage}
    \begin{minipage}{0.7\textwidth}
    A construction obstacle is placed outside the road to the right side, requiring no reaction of the ego vehicle. The scenario tests if models react to known cues even when they are placed outside the relevant regions. The in-distribution sample includes no scenario.
    \end{minipage}\\
    
    \item \textbf{OppositeConstruction}\\
    \begin{minipage}{0.3\textwidth}
    \includegraphics[width=0.98\textwidth]{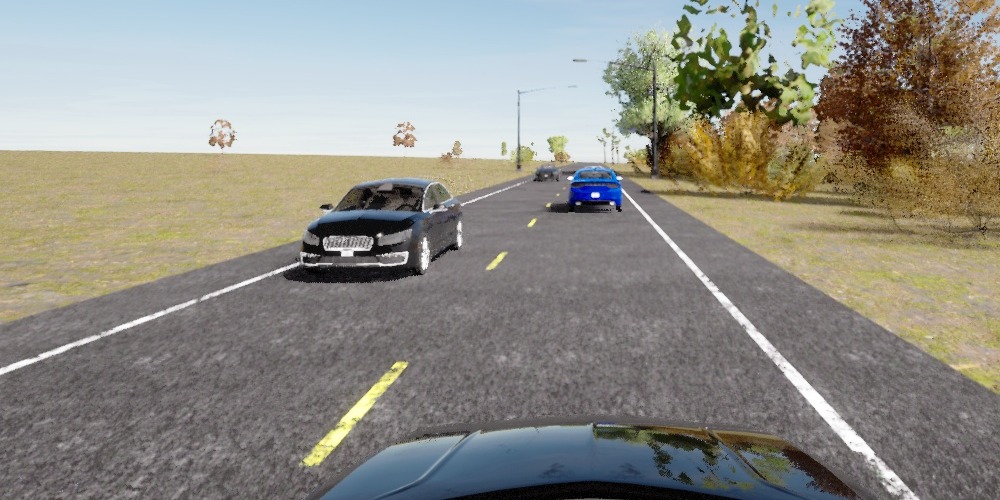}
    \includegraphics[width=0.98\textwidth]{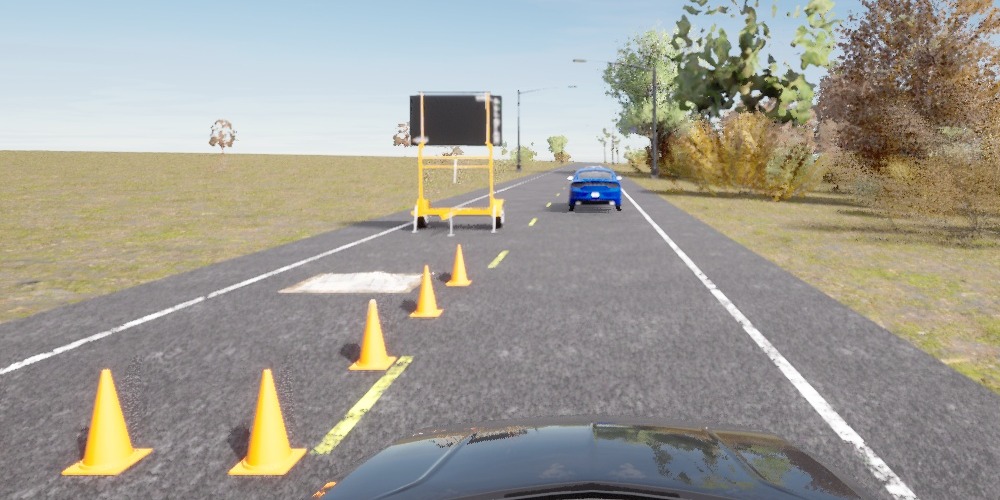}
    \end{minipage}
    \begin{minipage}{0.7\textwidth}
    A construction site is placed in the opposite lane, requiring no reaction from the ego vehicle, again testing overfitting to scenario structures. The in-distribution sample includes no scenario.
    \end{minipage}\\

    \item \textbf{ImageOnObject}\\
    \begin{minipage}{0.3\textwidth}
    \includegraphics[width=0.98\textwidth]{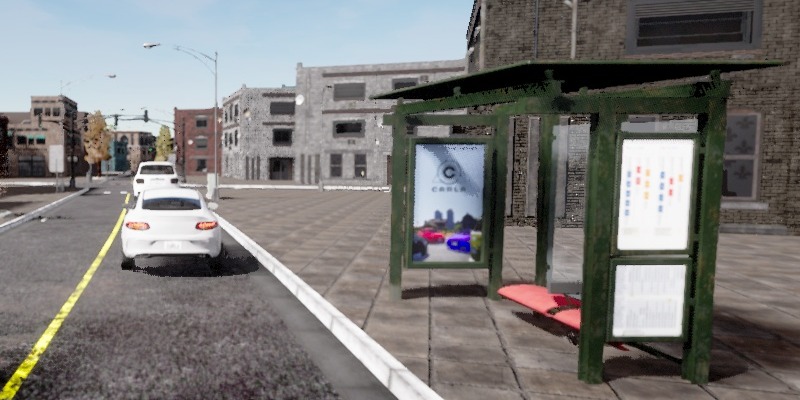}
    \includegraphics[width=0.98\textwidth]{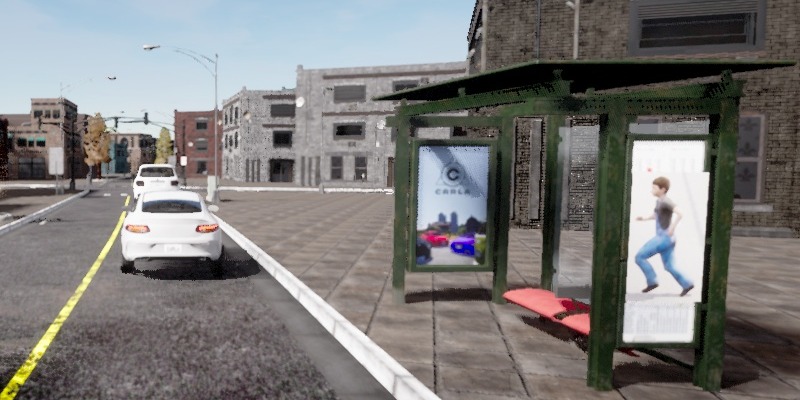}
    \end{minipage}
    \begin{minipage}{0.7\textwidth}
    A deceptive image is placed on an advertisement or a bus stop, the ego vehicle should not react to this influence. Images include a walking child at two scales and a red light, testing if models can differentiate between these printed images and real objects. The in-distribution scenario does not include an image.
    \end{minipage}\\

\newpage
    \item \textbf{PassableObstacles}\\
    \begin{minipage}{0.3\textwidth}
    \includegraphics[width=0.98\textwidth]{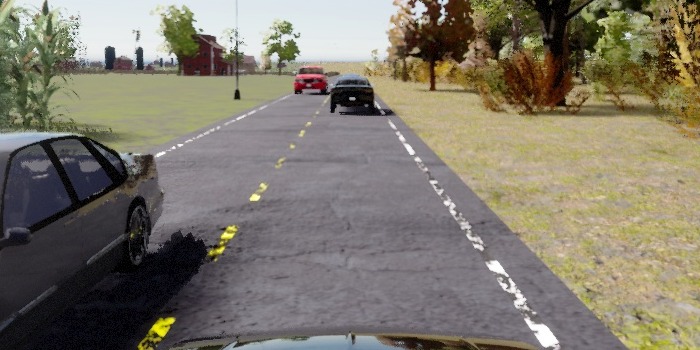}
    \includegraphics[width=0.98\textwidth]{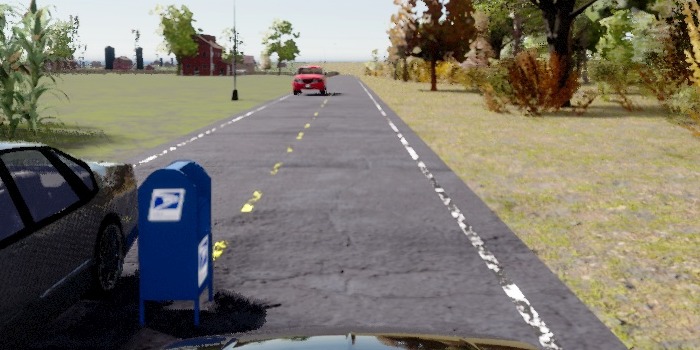}
    \end{minipage}
    \begin{minipage}{0.7\textwidth}
    Objects are placed on or near the road, allowing the vehicle to pass by maintaining its lane. This tests models' ability to disregard irrelevant objects that do not affect driving behavior. The in-distribution scenario includes no obstacles.
    \end{minipage}\\

    \item \textbf{PedestrianCrowd}\\
    \begin{minipage}{0.3\textwidth}
    \includegraphics[width=0.98\textwidth]{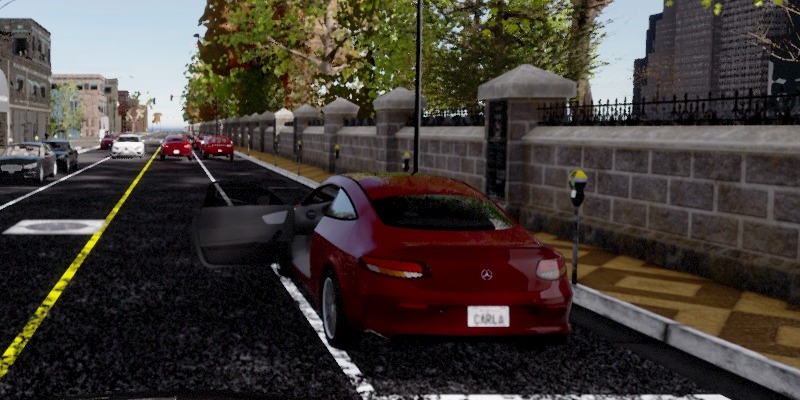}
    \includegraphics[width=0.98\textwidth]{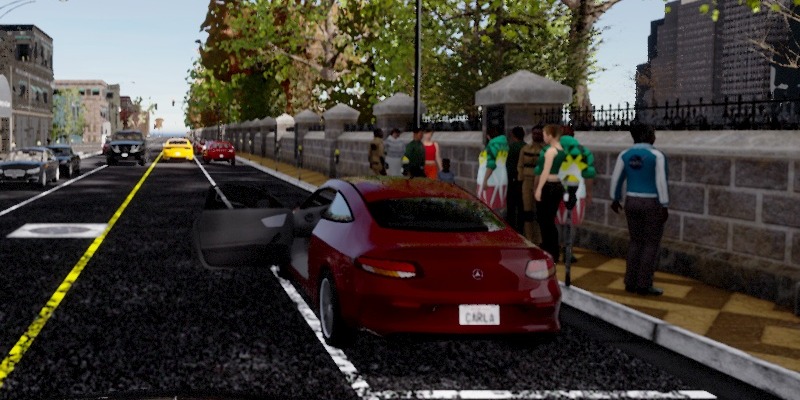}
    \end{minipage}
    \begin{minipage}{0.7\textwidth}
    A large number of pedestrians is standing on the sidewalk while the ego vehicle passes or performs a scenario. Since in CARLA v2, pedestrians are only present when relevant to a scenario, models may learn to react strongly to their presence. The in-distribution sample is defined by the same scenarios without any pedestrians.
    \end{minipage}\\

    \item \textbf{ConstructionPedestrian}\\
    \begin{minipage}{0.3\textwidth}
    \includegraphics[width=0.98\textwidth]{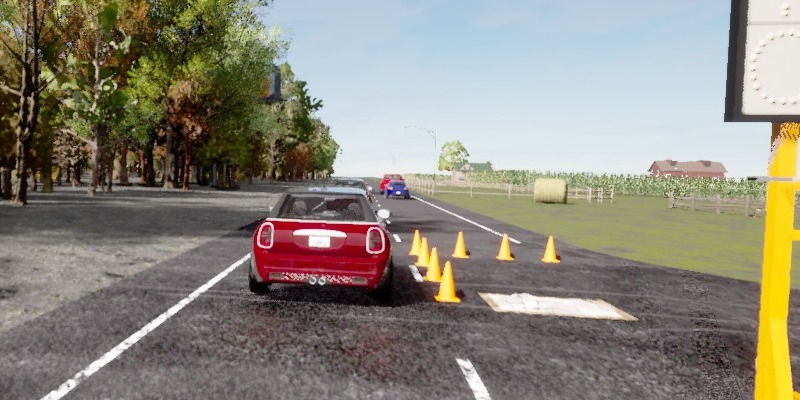}
    \includegraphics[width=0.98\textwidth]{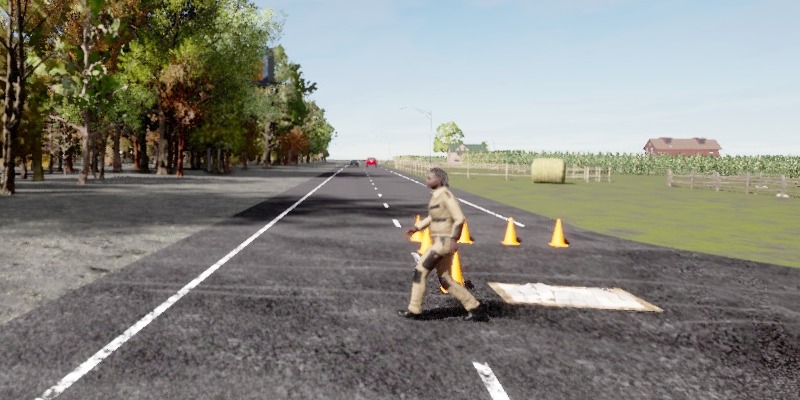}
    
    \end{minipage}
    \begin{minipage}{0.7\textwidth}
    While passing a construction site, a pedestrian crosses the road. This scenario requires the model to generalize to stop during the overtaking maneuver, which is not shown during training. The default \textit{ConstructionObstacle} without a pedestrian serves as the in-distribution sample.
    \end{minipage}\\

    \item \textbf{PedestriansOnRoad}\\
    \begin{minipage}{0.3\textwidth}
    \includegraphics[width=0.98\textwidth]{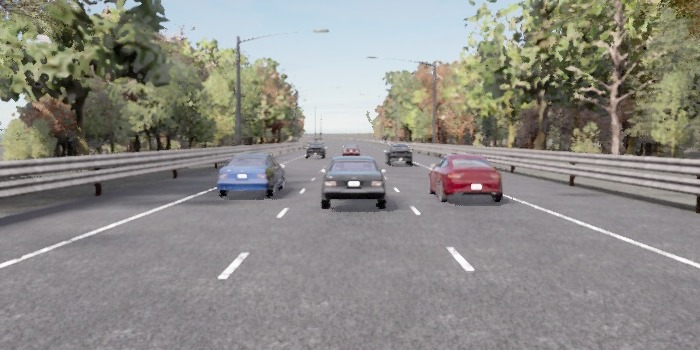}
    \includegraphics[width=0.98\textwidth]{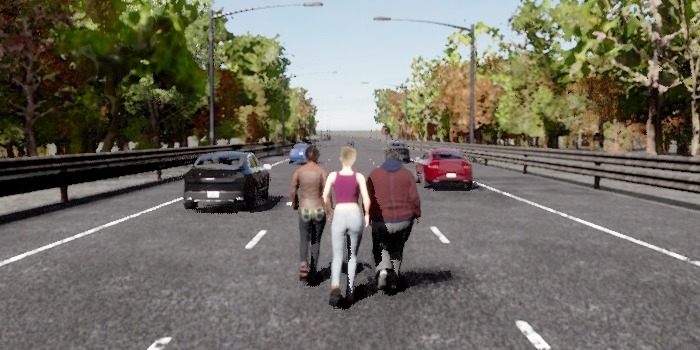}
    \end{minipage}
    \begin{minipage}{0.7\textwidth}
    Pedestrians are walking on the road in front of the ego vehicle, requiring deceleration or an evasive maneuver. This tests whether pedestrians are correctly identified and responded to in out-of-distribution scenarios. The in-distribution sample tests solving the underlying route without a scenario.
    \end{minipage}\\

\newpage 
    \item \textbf{FullyBlocked}\\
    \begin{minipage}{0.3\textwidth}
    \includegraphics[width=0.98\textwidth]{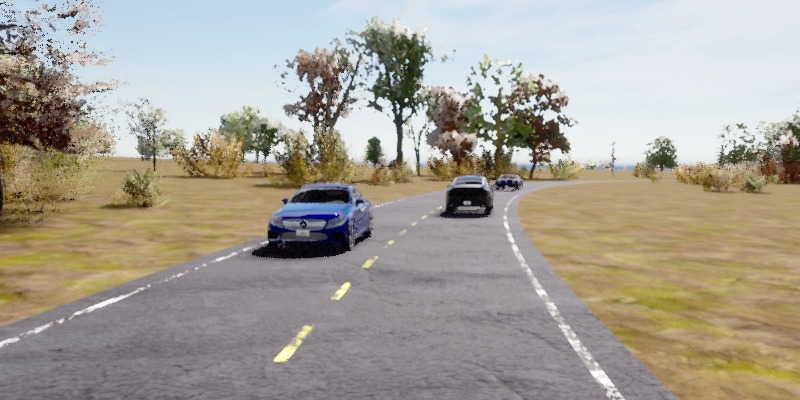}
    \includegraphics[width=0.98\textwidth]{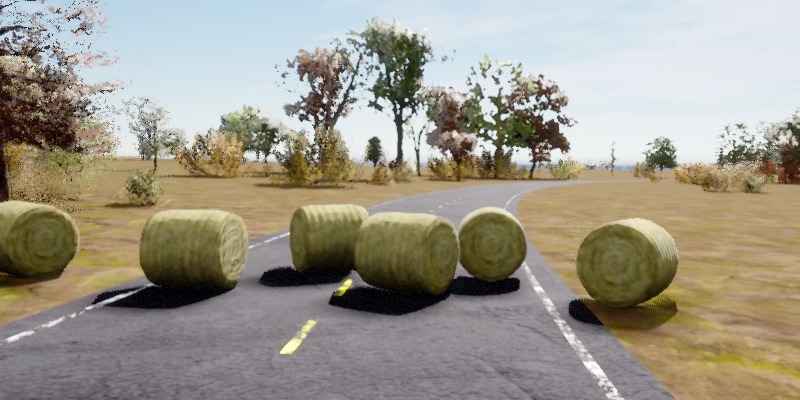}
    \end{minipage}
    \begin{minipage}{0.7\textwidth}
    An object blocks the entire road, forcing the ego vehicle to stop and wait 60 seconds, until the obstacle is removed and the vehicle can pass. While during training, only passable objects are shown, this scenario tests whether models generalize to stop and wait at obstacles. The in-distribution sample uses no scenario and evaluates a model's ability to complete the underlying road following task.
    \end{minipage}\\

    \item \textbf{Wall}\\
    \begin{minipage}{0.3\textwidth}
    \includegraphics[width=0.98\textwidth]{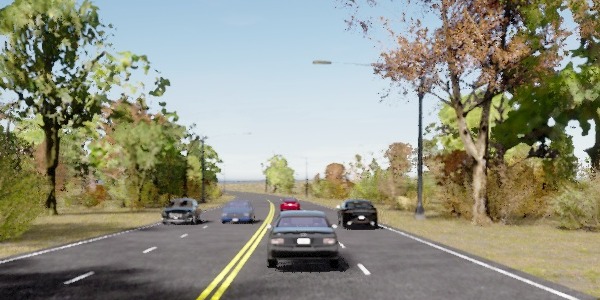}
    \includegraphics[width=0.98\textwidth]{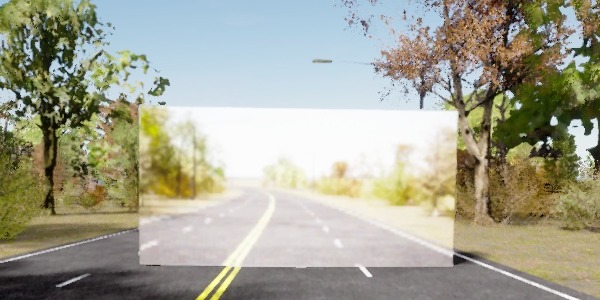}
    \end{minipage}
    \begin{minipage}{0.7\textwidth}
    A large-scale wall with a printed image is placed on the road, requiring the agent to wait for 60 seconds until the obstacle is removed. In addition to waiting at the object, this scenario introduces highly deceptive visuals. Fail2Drive includes one brick wall and three walls with images of roads. The in-distribution route is again defined without scenarios.
    \end{minipage}\\

\end{enumerate}

\section{Full results}
For completeness, we include numerical results for all models per generalization category in Table~\ref{tab:full_cat_results}.

\begin{table*}[h]
\centering
\begin{tabular}{l|cccccc}
\toprule

\textbf{Method} & \textbf{Visual-lon} & \textbf{Visual-lat} & \textbf{Behavior} & \textbf{Robustness}\\

\midrule

TCP~\cite{Wu2022NeurIPS} & 31.4 \small{(-10.3\%)} & 6.2 \small{(-1.4\%)} & 22.1 \small{(-30.6\%)} & 42.8 \small{(3.6\%)}\\
UniAD~\cite{hu2023_uniad} & 26.3 \small{(4.6\%)} & 13.0 \small{(84.2\%)} & 17.9 \small{(-67.9\%)} & 66.3 \small{(-4.7\%)}\\
Orion~\cite{Fu2025Orion} & 53.0 \small{(-9.6\%)} & 34.0 \small{(47.9\%)} & 35.4 \small{(-34.0\%)} & 66.0 \small{(-8.4\%)}\\
HiP-AD~\cite{tang2025hipadhierarchicalmultigranularityplanning} & 70.9 \small{(-5.6\%)} & 40.8 \small{(-27.1\%)} & 42.6 \small{(-46.6\%)} & 82.3 \small{(4.1\%)}\\
SimLingo~\cite{Renz2025cvpr} & 71.1 \small{(-9.0\%)} & 45.9 \small{(-32.2\%)} & 31.2 \small{(-64.2\%)} & 86.8 \small{(-5.9\%)}\\
\midrule
TF++~\cite{Jaeger2023ICCV} & 77.0 \small{(-7.8\%)} & 40.4 \small{(-30.6\%)} & 47.2 \small{(-43.9\%)} & 93.7 \small{(-2.2\%)}\\
\midrule\midrule
PlanT 2.0~\cite{gerstenecker2025plant2} & 86.6 \small{(-6.2\%)} & 36.2 \small{(-44.4\%)} & 36.7 \small{(-59.7\%)} & 82.9 \small{(-13.7\%)}\\

\end{tabular}
\caption{
\textbf{Categorized results on Fail2Drive.} Harmonic scores across generalization categories for all models.
\label{tab:full_cat_results}
}

\vspace{-4mm}

\end{table*}

\end{document}